\documentclass[lettersize,journal]{IEEEtran}
\usepackage{amsmath,amsfonts}
\usepackage{algorithmic}
\usepackage{algorithm}
\usepackage{adjustbox}
\usepackage{array,multirow}
\usepackage{textcomp}
\usepackage{stfloats}
\usepackage{url}
\usepackage{verbatim}
\usepackage[caption=false,font=footnotesize,labelfont=rm,textfont=rm]{subfig}
\usepackage{graphicx}
\usepackage{cite}
\begin{document}

\title{Muti-scale Graph Neural Network with\\ Signed-attention for Social Bot Detection:\\ A Frequency Perspective}

\author{Shuhao Shi, Kai Qiao, Zhengyan Wang, Jie Yang, Baojie Song, Jian Chen, Bin Yan
\thanks{This paper was produced by the IEEE Publication Technology Group. They are in Piscataway, NJ.}
\thanks{Manuscript received June 30, 2023; revised August 16, 2023.}}

\markboth{Journal of \LaTeX\ Class Files,~Vol.~14, No.~8, August~2023}%
{Shell \MakeLowercase{\textit{et al.}}: A Sample Article Using IEEEtran.cls for IEEE Journals}


\maketitle

\begin{abstract}
The presence of a large number of bots on social media has adverse effects. The graph neural network (GNN) can effectively leverage the social relationships between users and achieve excellent results in detecting bots. Recently, more and more GNN-based methods have been proposed for bot detection. However, the existing GNN-based bot detection methods only focus on low-frequency information and seldom consider high-frequency information, which limits the representation ability of the model. To address this issue, this paper proposes a Multi-scale with Signed-attention Graph Filter for social bot detection called MSGS. MSGS could effectively utilize both high and low-frequency information in the social graph. Specifically, MSGS utilizes a multi-scale structure to produce representation vectors at different scales. These representations are then combined using a signed-attention mechanism. Finally, multi-scale representations via MLP after polymerization to produce the final result. We analyze the frequency response and demonstrate that MSGS is a more flexible and expressive adaptive graph filter. MSGS can effectively utilize high-frequency information to alleviate the over-smoothing problem of deep GNNs. Experimental results on real-world datasets demonstrate that our method achieves better performance compared with several state-of-the-art social bot detection methods.
\end{abstract}


\section{Introduction}
\IEEEPARstart{S}{ocial} media have become an indispensable part of people's daily lives. However, the existence of automated accounts, also known as social bots, has brought many problems to social media. These bots have been employed to disseminate false information, manipulate elections, and deceive users, resulting in negative societal consequences~\cite{article01,article02,article03}. Effectively detecting bots on social media plays an essential role in protecting user interests and ensuring stable platform operation. Therefore, the accurate detection of bots on social media platforms is becoming increasingly crucial.

Graph neural networks (GNNs) have emerged as powerful tools for processing non-Euclidean data, where entities are represented as nodes and relationships as edges in a graph. Leveraging the inherent graph structure, GNNs enable convolutions on the graph data, facilitating effective utilization of the relationships between entities. GNNs have demonstrated impressive performance in the field of social account detection. Building upon GNN-based approaches~\cite{article15,article16,article36}, researchers have formulated the social bot detection task as a node classification problem. Alhosseini et al.~\cite{article13} were pioneers in utilizing graph convolutional neural networks (GCNs)~\cite{article17} to detect bots, effectively leveraging the graph structure and relationships among Twitter accounts. Subsequent investigations have focused on exploring multiple relationships within social graphs. For instance, Feng et al.~\cite{article15} introduced the Relational Graph Convolutional Network (RGCN)~\cite{article29} for Twitter social bot detection, enabling the integration of multiple social relationships between accounts. Additionally, Shi et al.~\cite{article16} proposed a graph learning data augmentation technique to address the challenges of class-imbalance in socail bot detection.

\begin{figure}[htbp]
	\centering
	\begin{minipage}{1.0\linewidth}
		\centering
		\includegraphics[width=1.0\linewidth]{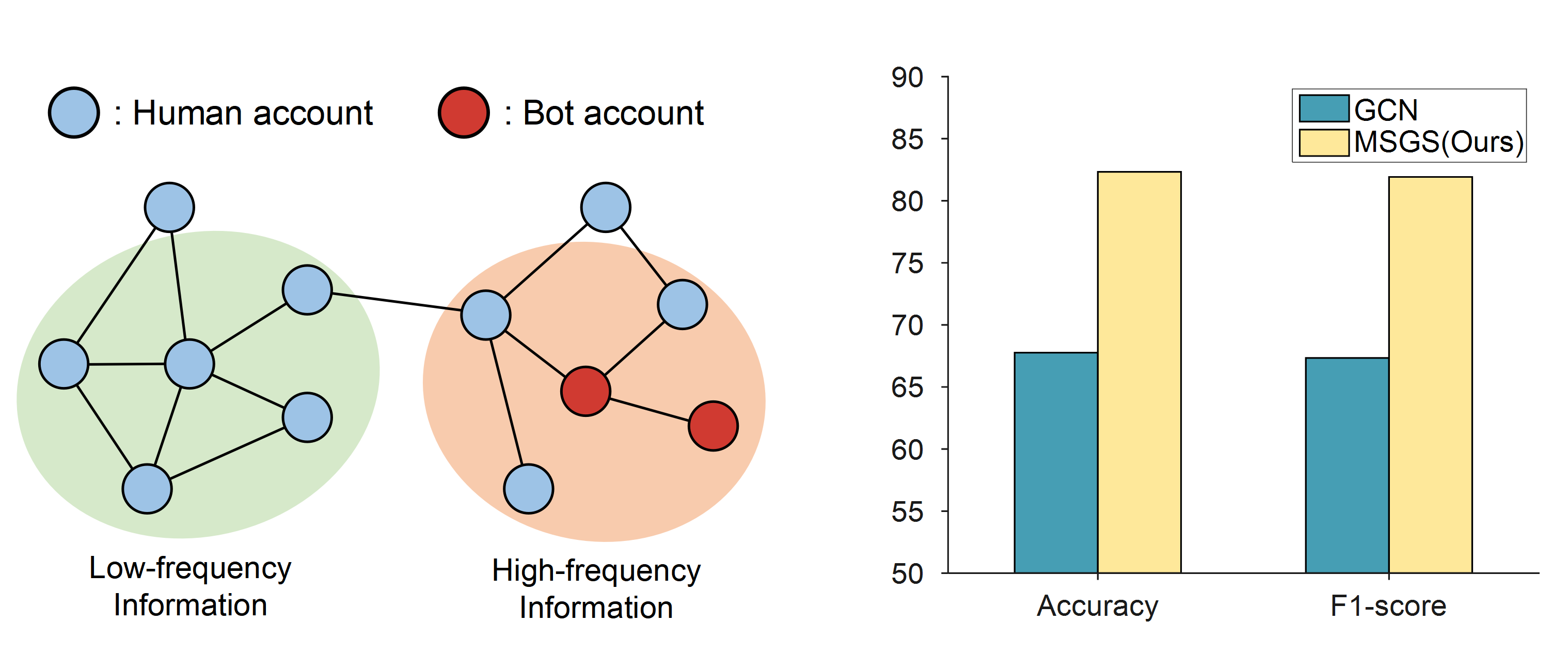}
		\caption{Left: An illustration of graph in social bot detection. Accounts have different features or common features from the neighbors indicate high-frequency and low-frequency information, respectively. Right: The performance of GCN and our proposed MSGS on the MGTAB dataset.}
		\label{fig:low_high}
	\end{minipage}
\end{figure}

Existing GNNs mainly apply fixed filters for the convolution operation, these models assuming that nodes tend to share common features with their neighbors (low-frequency information)~\cite{article04,article05,article06}. However, this assumption may be weakened in networks containing anomalies, since anomalies tend to have different features from the neighbors (high-frequency signals)~\cite{article07,article09}. As shown in Fig.~\ref{fig:low_high}, using low-frequency information alone is insufficient in social bot detection. In view of the shortcoming that GNN cannot effectively utilize the high-frequency information in the user network, we designed a more flexible GNN structure that can adapt to learn the low-frequency and high-frequency information.

Our proposed framework pioneers the exploration of high-frequency signals in social bot detection, harnessing the power of GNNs. We introduce a novel GNN framework called MSGS, which adeptly captures the varying significance of different frequency components for node representation learning. At the core of this framework lies a simple yet elegant trainable filter, constructed through a multi-scale architecture and symbol attention mechanism that across multiple layers. By employing multi-scale features, we train a graph filter that intelligently exploits low-frequency and high-frequency information. Our extensive experimental results demonstrate the remarkable performance enhancement of GNNs on various benchmark datasets for social bot detection achieved by our proposed framework. The main contributions of our work are as follows:

\begin{itemize}
\item We are the first to analyze the high-frequency information in social bot detection and highlight the shortcomings of traditional GNNs in effectively utilizing it.
\item Our proposed MSGS combines multi-scale architecture and signed-attention mechanism, enabling adaptive learning of the frequency response of the graph filter, thereby effectively leveraging both low-frequency and high-frequency information in social bot detection.
\item Extensive experiments on real-world social bot detection datasets establish that MSGS outperforms other leading methods, including multi-scale GNNs and spectral GNNs.
\end{itemize}

\section{Preliminaries}
In this section, we define some notations and used them throughout this paper. Let $\mathcal{G}=(\mathcal{V}, \mathcal{E})$ denote the user networks graph, where $\mathcal{V}=\left\{v_{1}, \cdots, v_{N}\right\}$ is the set of vertices with $|\mathcal{V}|=N$ and $\mathcal{E}$ is the set of edges. The adjacency matrix is defined as $\mathbf{A} \in\{0,1\}^{N \times N}$, and $\mathbf{A}_{i, j}=1$ if and only if there is a edge between $v_{i}$ and $v_{j}$. $\mathbf{D} \in \mathbb{R}^{N \times N}$ is the degree matrix of $\mathbf{A}$. $\mathbf{D}=\operatorname{diag}\left\{d_{1}, d_{2}, \ldots, d_{N}\right\}$ and $d_{i}=\sum_{j} \mathbf{A}_{i j}$. Let $\mathcal{N}_{i}$ represents the neighborhood of node $v_{i}$. The feature matrix is represent as $\mathbf{X} \in \mathbb{R}^{N \times M}$, where each node $v$ is associated with a $M$ dimensional feature vector $\mathbf{X}_{v}$.

\subsection{Graph Fourier Transform}
\noindent \textbf{Theorem 1} (Convolution theorem) The Fourier transform of the convolution of functions is the product of the Fourier transforms of functions. For functions $f$ and $g$, $\mathcal{F}\{\cdot\}$ and $\mathcal{F}^{-1}\{\cdot\}$ represent Fourier transform and Inverse Fourier transform respectively, then $f * g=\mathcal{F}^{-1}\{\mathcal{F}\{f\} \cdot \mathcal{F}\{g\}\}$. The proof of \textbf{Theorem 1} is provided in Appendix.

The graph spectral analysis relies on the spectral decomposition of graph Laplacians. Ordinary forms of Laplacian matrix is defined as $\mathbf{L}=\mathbf{D}-\mathbf{A}$, The normalized form of Laplace matrix is defined as $\mathbf{L}_{s v m}=\mathbf{I}-\mathbf{D}^{-1 / 2} \mathbf{A} \mathbf{D}^{-1 / 2}$. The the random walk normalized form of Laplace matrix is defined as $\mathbf{L}_{r w}=\mathbf{D}^{-1} \mathbf{L}=\mathbf{I}-\mathbf{D}^{-1} \mathbf{A}$. In this paper, we only analyze the normalized graph Laplacian matrix $\mathbf{L}_{sym}$. The analysis results can be easily extended to other Laplacian matrices. The purpose of defining the Laplacian operator is to find the basis for Fourier transforms. The Fourier basis on the graph is made up of the eigenvectors of the $\mathbf{L}$, $\mathbf{U}=\left[\mathbf{u}_{1} \ldots \mathbf{u}_{n}\right]$. The eigenvalue decomposition of the Laplace matrix can be expressed as $\mathbf{L}=\mathbf{U} \boldsymbol{\Lambda} \mathbf{U}^{T}$, where $\boldsymbol{\Lambda}=\operatorname{diag}\left(\left[\lambda_{1}, \lambda_{2}, \cdots, \lambda_{n}\right]\right)$ is a diagonal matrix of $\mathbf{L}$’s eigenvalues, $\lambda_{l} \in[0,2]$ and $1 \leq l \leq N$. Assuming $\lambda_{1} \leq \lambda_{2} \leq \ldots \leq \lambda_{N}$, $\lambda_{1}$ and $\lambda_{N}$ correspond to the lowest and the highest frequency of the graph.

\subsection{Graph Spectral Filtering}
Signal filtering is a crucial operation in signal processing. It extracts or enhances the required frequency components in the input signal and filters or attenuates some unwanted frequency components. According to \textbf{Theorem 1}, the signal is first transformed into the frequency domain, multiplied element-by-element in the frequency domain, and finally transformed back into the time domain. A graph signal $\mathbf{x}$ with filter $f$ of the eigenvalues can be defined as follows:

\begin{equation}
\label{equ:1}
\mathbf{H}=f * \mathbf{x}=\mathbf{U}\left(\left(\mathbf{U}^{T} f\right) \odot\left(\mathbf{U}^{T} \mathbf{x}\right)\right),
\end{equation}

\noindent where $\hat{\mathbf{x}}=\mathbf{U}^{\top} \mathbf{x}$ denotes the graph Fourier transform, and $\mathbf{x}=\mathbf{U} \hat{\mathbf{x}}$ denotes Inverse Fourier transform. $\bigodot$ denotes element-wise multiplication. $\mathbf{U}^{T} f=\left[g\left(\lambda_{1}\right), g\left(\lambda_{2}\right), \ldots, g\left(\lambda_{n}\right)\right]^{T}$ is called the convolution filter in the frequency domain. Define $g_{\theta}(\Lambda)=\operatorname{diag}\left(\left[g\left(\lambda_{1}\right), g\left(\lambda_{2}\right), \ldots, g\left(\lambda_{n}\right)\right]\right)$, and $\theta$ is the learnable convolution kernel parameter, then:

\begin{equation}
\label{equ:2}
\mathbf{H}=f * \mathbf{x}=\mathbf{U} g_{\theta} \mathbf{U}^{T} \mathbf{x}.
\end{equation}

The computational complexity of graph convolution is high because of the high cost of eigenvalue decomposition for graph’s Laplacian. To overcome the disadvantage of having a large convolution kernel, ChebNet approximates the parameterized frequency response function with a $K$-order polynomial $g_{\theta}=\sum_{k=0}^{K} \theta_{i} \Lambda^{i}$, then:

\begin{equation}
\label{equ:3}
\mathbf{x} * \mathbf{g} \approx \mathbf{U}\left(\sum_{i=0}^{K} \theta_{i} \mathbf{\Lambda}^{i}\right) \mathbf{U}^{T} \mathbf{x}=\sum_{i=0}^{K} \theta_{i} \mathbf{L}_{\mathrm{n}}^{i} \mathbf{x}.
\end{equation}

Thomas et al. [GCN] proposed a simpler graph convolution which approximates first-order Chebyshev graph convolution. Specifically, let $\theta_{0}=2 \theta$, $\theta_{1}=-\theta$, $\theta_{k>1}=0$:

\begin{equation}
\label{equ:4}
\mathbf{x} * \mathbf{g} \approx \theta\left(2 \mathbf{I}-\mathbf{L}_{\mathrm{n}}\right) \mathbf{x}=\theta(\mathbf{I}+\mathbf{D}^{-1 / 2} \mathbf{A} \mathbf{D}^{-1 / 2}) \mathbf{x}.
\end{equation}

\noindent \textbf{Theorem 2} (Over-smoothing) For any fixed low-pass graph filters defined over $\mathbf{L}^{sym}$, given a graph signal $\mathbf{x}$, suppose we convolve $\mathbf{x}$ with the graph filter. If the number of layers in the GNN is large enough, the over-smoothing issue becomes inevitable. The proof of \textbf{Theorem 2} is provided in Appendix.

\section{The Proposed Method}
The use of fixed low-pass filters in GCN and other GNNs largely limits the expressive power of GNNs, thereby affecting their performance. The novelty of our method lies in the multi-scale and signed attention. Through the use of directional attention and coefficients $\boldsymbol{\gamma}^{(0)}, \boldsymbol{\gamma}^{(1)}, \ldots, \boldsymbol{\gamma}^{(K)}$ of different scale channels, we learn the filtering function. MSGS works well universally by effectively utilizing low-frequency and high-frequency information through learning frequency hyperparameters to change the frequency spectrum of the graph filter.

\begin{figure}[htbp]
	\centering
	\begin{minipage}{0.975\linewidth}
		\centering
		\includegraphics[width=0.95\linewidth]{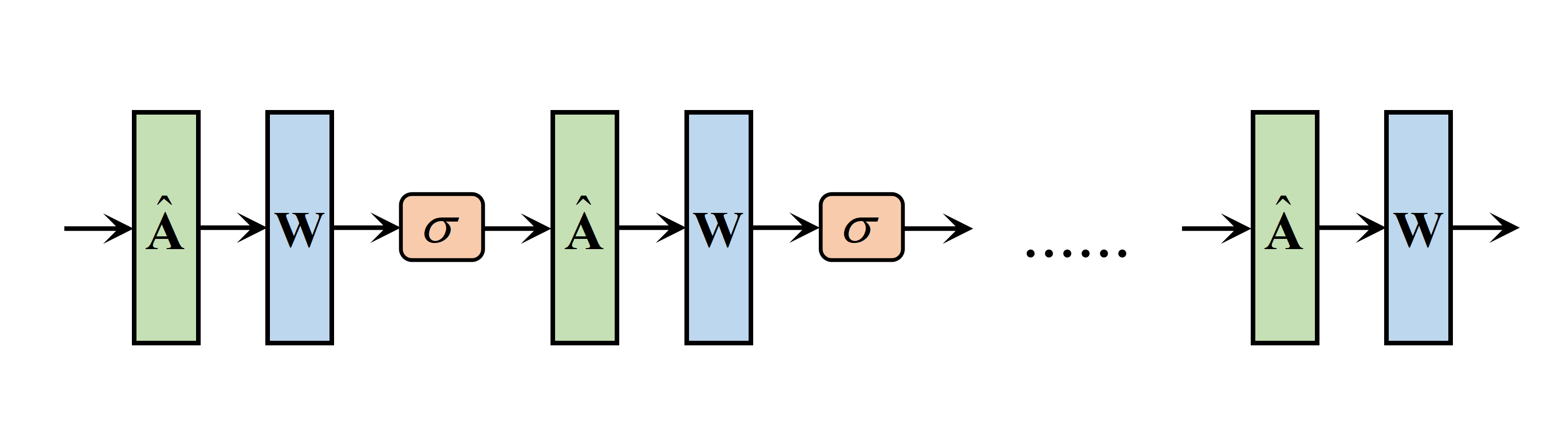}
		\caption{Architecture of common GNNs.}
		\label{fig:GCN_architect}
	\end{minipage}

	\begin{minipage}{0.975\linewidth}
		\centering
		\includegraphics[width=0.95\linewidth]{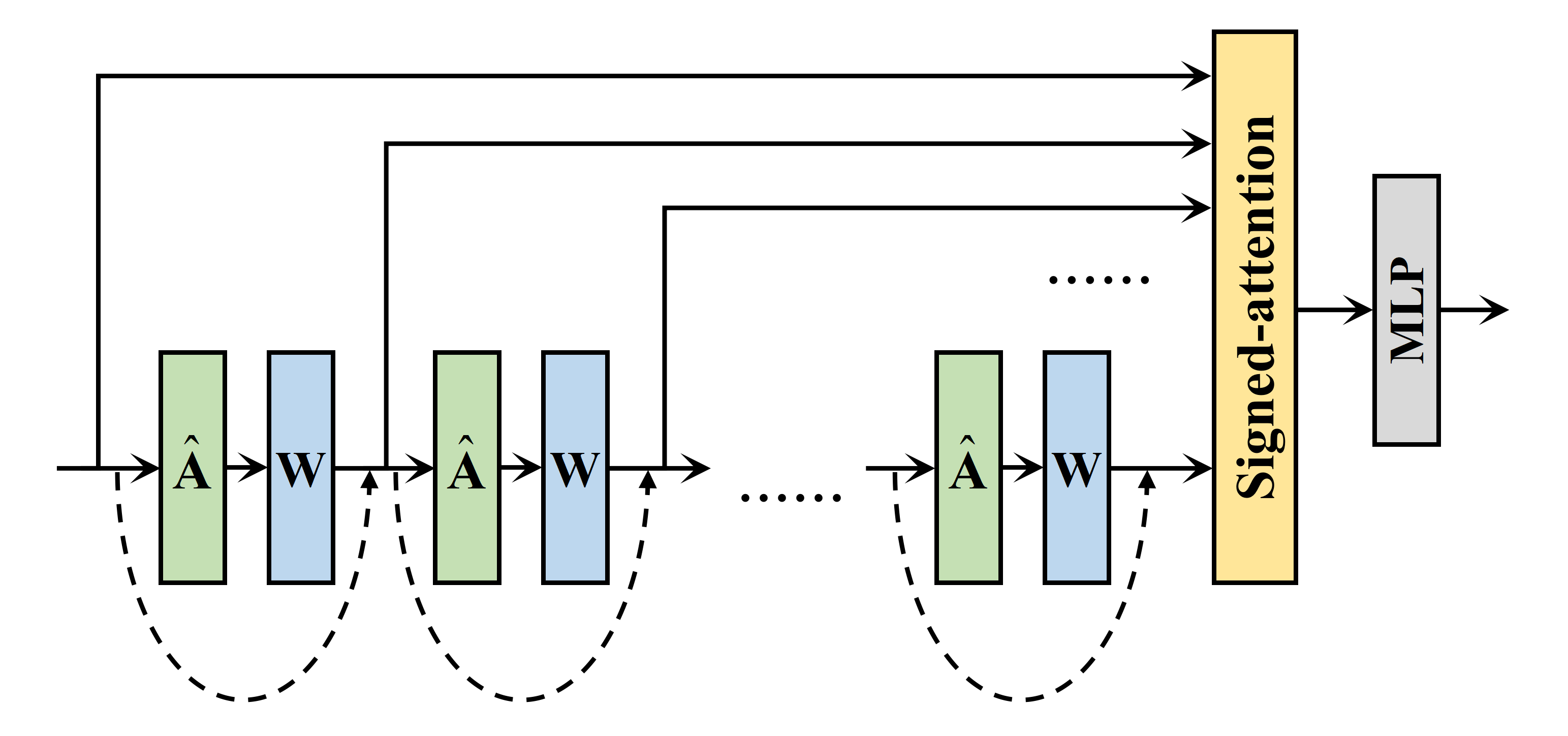}
		\caption{Architecture of our proposed MSGF.}
		\label{fig:MSGS_architect}
	\end{minipage}
\end{figure}

\subsection{Muti-scale Architecture}
\label{sec:3.1}
\noindent \textbf{Proposition 1.} Most existing GNN models, such as GCN, employ a fixed low-pass filter. As a result, after passing through a GNN, the node representations become similar.
Assume that $\left(v_{i}, v_{j}\right)$ is a pair of connected nodes, $\mathbf{x}_{i}$ and $\mathbf{x}_{j}$ are the node features. $\mathcal{D}_{i, j}$ represents the distance between nodes $v_{i}$ and $v_{j}$. The original distance of representations is $\mathcal{D}_{i, j}=\left\|\mathbf{x}_{i}-\mathbf{x}_{j}\right\|_{2}$. The filter used in GCN is $\mathbf{I}+\mathbf{D}^{-1 / 2} \mathbf{A} \mathbf{D}^{-1 / 2}$. Subject to $d_{i} \approx {} d_{j} \approx d$, the distance of representations learned after neighborhood aggregation is:

\begin{equation}
\label{equ:5}
\tilde{\mathcal{D}}_{i, j} \approx\left\|(\mathbf{x}_{i}+\frac{\mathbf{x}_{j}}{d_{j}})-(\mathbf{x}_{j}+\frac{\mathbf{x}_{i}}{d_{i}})\right\|_{2} \approx\left\|1-\frac{1}{d}\right\|_{2}<\mathcal{D}_{i, j}
\end{equation}

After neighborhood aggregation by GNN, the distance between node representations decreases.
Although different GNN models use different $f$ in Equ.~(\ref{equ:2}), GCN and many subsequent models use a fixed low-pass filter for graph convolution, leading to similar node representations. According to \textbf{Theorem 2}, when the number of model layers is too deep, it will lead to the overs-smoothing issue in GNN. When using multiple GNN layers for learning, the task performance decline significantly.
To improve the ability of GNN models to utilize the information at different frequencies, we propose a multi-scale graph learning framework. Specifically, the feature embedding of the $l$-th layer of the GCN model is defined as follows:

\begin{equation}
\label{equ:6}
\mathbf{H}^{(l)}=\sigma(\hat{\mathbf{A}} \mathbf{H}^{(l-1)} \mathbf{W}^{(l)}),
\end{equation}

\noindent where $\mathbf{W}^{(l)}$ is a learnable parameter matrix and $l \geq 1$, $\mathbf{H}^{(0)}=\mathbf{X} \mathbf{W}^{(0)}$. $\sigma(\cdot)$ is the activation function. $\mathbf{H}^{(1)}$ represents the feature embedding obtained by passing $l$-layer of graph convolution. $\mathbf{H}^{(0)}, \mathbf{H}^{(1)}, \ldots, \mathbf{H}^{(K)}$ are feature embeddings obtained at different scales. Let $\tilde{\mathbf{H}}^{(l)}$ denote the feature embedding obtained after neighborhood aggregation, $\tilde{\mathbf{H}}^{(l)}=\hat{\mathbf{A}} \mathbf{H}^{(l)}$. We retain both the embeddings before and after feature propagation:

\begin{equation}
\label{equ:7}
\mathbf{Z}^{(l)}=(\alpha^{(l)}-\beta^{(l)}) \mathbf{H}^{(l)}+\beta^{(l)} \tilde{\mathbf{H}}^{(l)}.
\end{equation}

The calculation of $\alpha^{(l)}$ and $\beta^{(l)}$ is detailed in Section \ref{sec:3.2}. $\mathbf{P}$ contains adaptive filters with $K+1$ different scales, shown in Equ.~(\ref{equ:8}). The coefficients $\boldsymbol{\Gamma}^{(0)}, \boldsymbol{\Gamma}^{(1)}, \ldots, \boldsymbol{\Gamma}^{(K)}$ are calculated through scale-level attention mechanism, see Section \ref{sec:3.2} for details.

\begin{equation}
\label{equ:8}
\mathbf{P}=\sum_{k=0}^{K} \boldsymbol{\Gamma}^{(k)} \cdot \mathbf{Z}^{(k)}.
\end{equation}

\subsection{Signed-attention Mechanism}
\label{sec:3.2}
\noindent \textbf{Node-level attention mechanism}
In Equ.~(\ref{equ:7}), $\alpha^{(l)} \in(0,1]$ and $\beta^{(l)} \in(-1,1)$. $\alpha^{(l)}-\beta^{(l)}$ controls the proportion of preserved original embedded features, while $\beta^{(l)}$ is the coefficient of the aggregated neighborhood features.

\noindent \textbf{Proposition 2.} The graph filter $g$: $\mathbf{Z}^{(K)}=\left(\alpha^{(K)}-\beta^{(K)}\right) \mathbf{H}^{(K)}+\beta^{(K)} \tilde{\mathbf{H}}^{(K)}$ is an adaptive filter that can be adjusted to a low-pass or high-pass filter depending on the changes of $\alpha^{(K)}$ and $\beta^{(K)}$.
The filter used in $g$ is $\alpha^{(K)} \mathbf{I}+\beta^{(K)} \mathbf{D}^{-1 / 2} \mathbf{A} \mathbf{D}^{-1 / 2}$, then:

\begin{equation}
\begin{split}
\label{equ:9}
\tilde{\mathcal{D}}_{i, j} &\approx\left\|\left(\alpha^{(K)} \mathbf{x}_{i}+\frac{\beta^{(K)} \mathbf{x}_{j}}{d_{j}}\right)-\left(\alpha^{(K)} \mathbf{x}_{j}+\frac{\beta^{(K)} \mathbf{x}_{i}}{d_{i}}\right)\right\|_{2} \\
&\approx \alpha^{(K)}\left\|1-\frac{\beta^{(K)}}{d}\right\|_{2} \mathcal{D}_{i, j}\left(\text { s.t. }  \mathrm{du} \approx  \mathrm{dv} \approx \mathrm{d}\right).
\end{split}
\end{equation}

When $\alpha^{(K)}\left\|1-\frac{\beta^{(K)}}{d}\right\|_{2}<1$, $\tilde{\mathcal{D}}_{i, j}<\mathcal{D}_{i, j}$, $g$ is a low-pass filter. When $\alpha^{(K)}\left\|1-\frac{\beta^{(K)}}{d}\right\|_{2}<1$, $\tilde{\mathcal{D}}_{i, j}<\mathcal{D}_{i, j}$, $g$ becomes a high-pass filter. High-pass filtering makes the representations become discriminative.
The proper design of $\alpha^{(K)}$ and $\beta^{(K)}$ requires knowing whether the information in the graph is high frequency or low frequency. However, we usually do not know the frequency distribution of the graph signal. Therefore, we propose a shared adaptive mechanism to calculate node-specific frequency coefficients $\alpha_{i}^{(K)}$ and $\beta_{i, j}^{(K)}$:

\begin{equation}
\label{equ:10}
\alpha_{i}^{(K)}=\sigma(\mathbf{g}_{\alpha}^{(K)}\left[\tilde{\mathbf{h}}_{i}^{(K)}-\mathbf{h}_{i}^{(K)}\right]),
\end{equation}

\begin{equation}
\label{equ:11}
\beta_{i, j}^{(K)}=\sigma((\mathbf{g}_{\beta}^{(K)})^{T}\left[\mathbf{h}_{i}^{(K)} \| \mathbf{h}_{j}^{(K)}\right]),
\end{equation}

\noindent where $\mathbf{g}_{\alpha}^{(K)}$ and $\mathbf{g}_{\beta}^{(K)}$ are shared attention vectors, the more similar $\tilde{\mathbf{h}}_{i}^{(K)}$ and $\mathbf{h}_{i}^{(K)}$ is, the smaller $\alpha_{i}^{(K)}$ tend to be.

\noindent \textbf{Scale-level attention mechanism} Calculate the attention coefficients $(\boldsymbol{\Gamma}^{(0)}, \boldsymbol{\Gamma}^{(1)}, \ldots, \boldsymbol{\Gamma}^{(K)})$ of the multi-scale feature embeddings through a signed-attention mechanism:

\begin{equation}
\label{equ:12}
(\boldsymbol{\Gamma}^{(0)}, \boldsymbol{\Gamma}^{(1)}, \ldots, \boldsymbol{\Gamma}^{(K)})=\operatorname{att}(\mathbf{Z}^{(0)}, \mathbf{Z}^{(1)}, \ldots, \mathbf{Z}^{(K)})
\end{equation}

where $\boldsymbol{\alpha}^{(k)} \in R^{N \times 1}$ represents the attention value vector of embeddings $\mathbf{Z}^{(K)}$ for $N$ node, $0 \leq k \leq K$. For node $v_i$, its feature embedding in the $(k+1)$th scale is $\mathbf{z}_{i}^{(k)}$, which represents the $i$-th row of $\mathbf{Z}^{(k)}$, $(\mathbf{Z}^{(k)})^{T}=(\mathbf{z}_{1}^{(k)}, \mathbf{z}_{2}^{(k)}, \ldots, \mathbf{z}_{N}^{(k)})$. The feature embedding is nonlinearly transformed and then attention values are obtained through a shared attention vector $\mathbf{q}$:

\begin{equation}
\gamma_{k, i}=\mathbf{q}^{T} \cdot \tanh (\mathbf{W}_{k} \cdot(\mathbf{z}_{i}^{(k)})^{T}).
\end{equation}

$\boldsymbol{\Gamma}^{(k)}=\left[\gamma_{k, i}\right]$, $0<i \leq N$. Once all the coefficients are computed, we can obtain the final embedding $\mathbf{P}$ according to Equ.~(\ref{equ:8}). Then, we use the output embedding for semi-supervised node classification with a linear transformation and a softmax function:

\begin{equation}
\hat{\mathbf{Y}}_{i}=\operatorname{softmax}\left(\mathbf{W} \cdot \mathbf{P}_{i}+\mathbf{b}\right),
\end{equation}

\noindent where $\mathbf{W}$ and $\mathbf{b}$ are learnable parameters, softmax is actually a normalizer across all classes. Suppose the training set is $\mathrm{V}_{L}$, for each $v_{n} \in \mathrm{V}_{L}$ the real label is $\mathbf{y}_{n}$ and the predicted label is $\tilde{\mathbf{y}}_{n}$. In this paper, we employs Cross-Entropy loss to measure the supervised loss between the real and predicted labels. The loss function is as follows:

\begin{equation}
\mathcal{L}=-\sum_{v_{n} \in \mathrm{V}_{L}} \operatorname{loss}\left(\mathbf{y}_{n}, \tilde{\mathbf{y}}_{n}\right).
\end{equation}

\section{Theoretical Analysis}
\subsection{Spectral Analysis for GCN}

\begin{figure}[htbp]
	\centering
    \subfloat[]{
	\begin{minipage}{0.475\linewidth}
		\centering
		\includegraphics[width=1\linewidth]{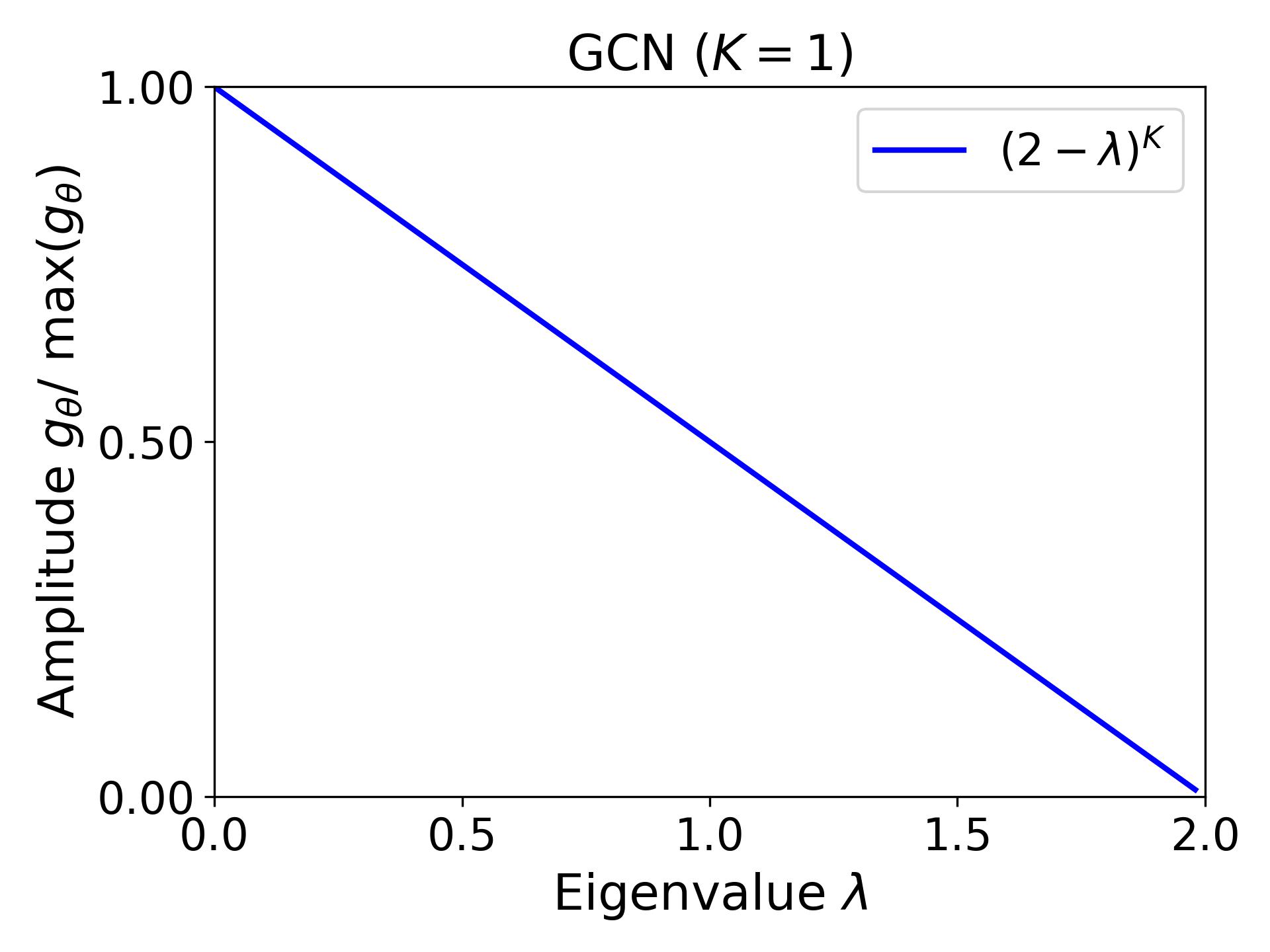}
	\end{minipage}
    }
    \subfloat[]{
	\begin{minipage}{0.475\linewidth}
		\centering
		\includegraphics[width=1\linewidth]{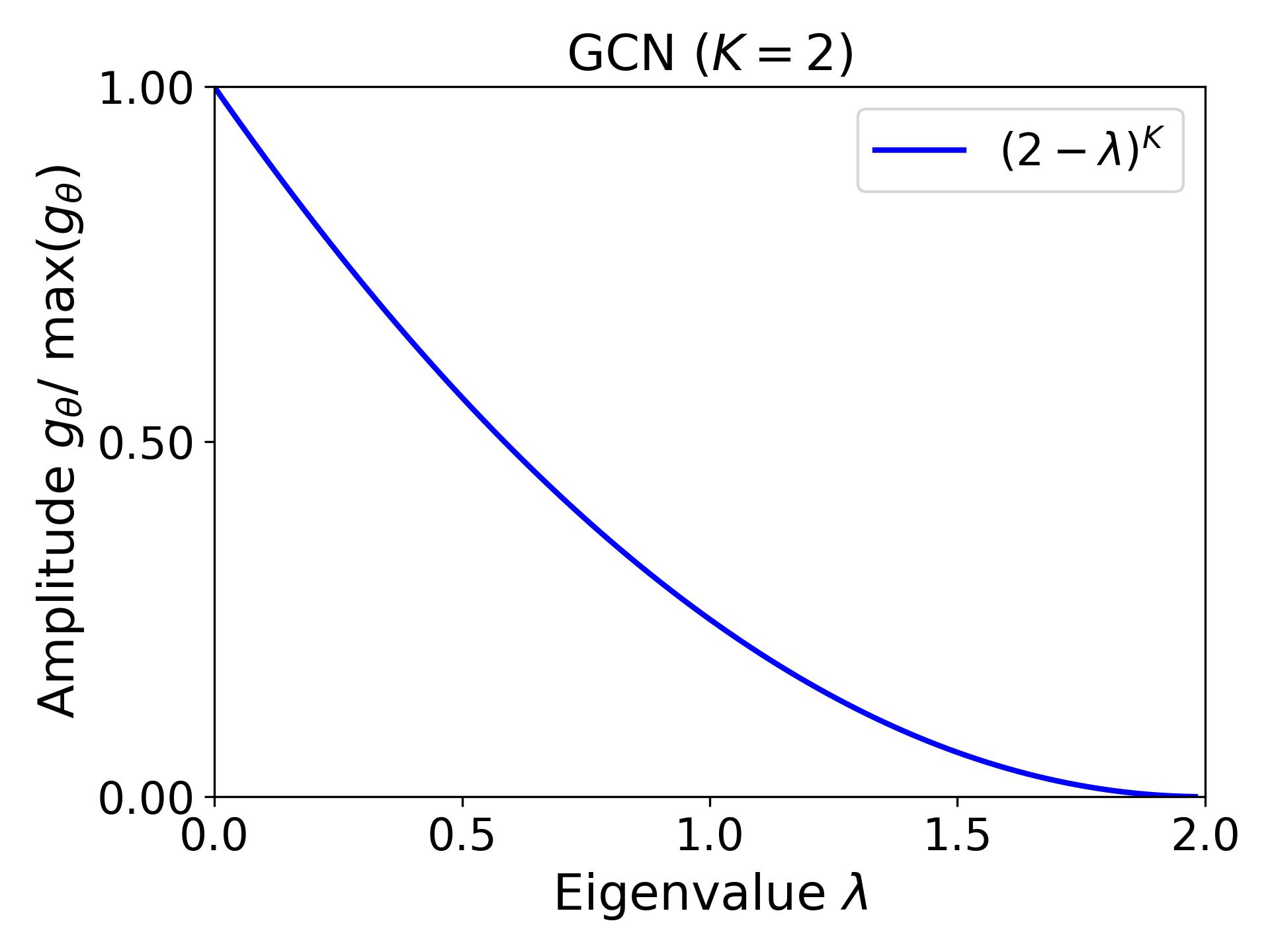}
	\end{minipage}
    }

	\subfloat[]{
	\begin{minipage}{0.475\linewidth}
		\centering
		\includegraphics[width=1\linewidth]{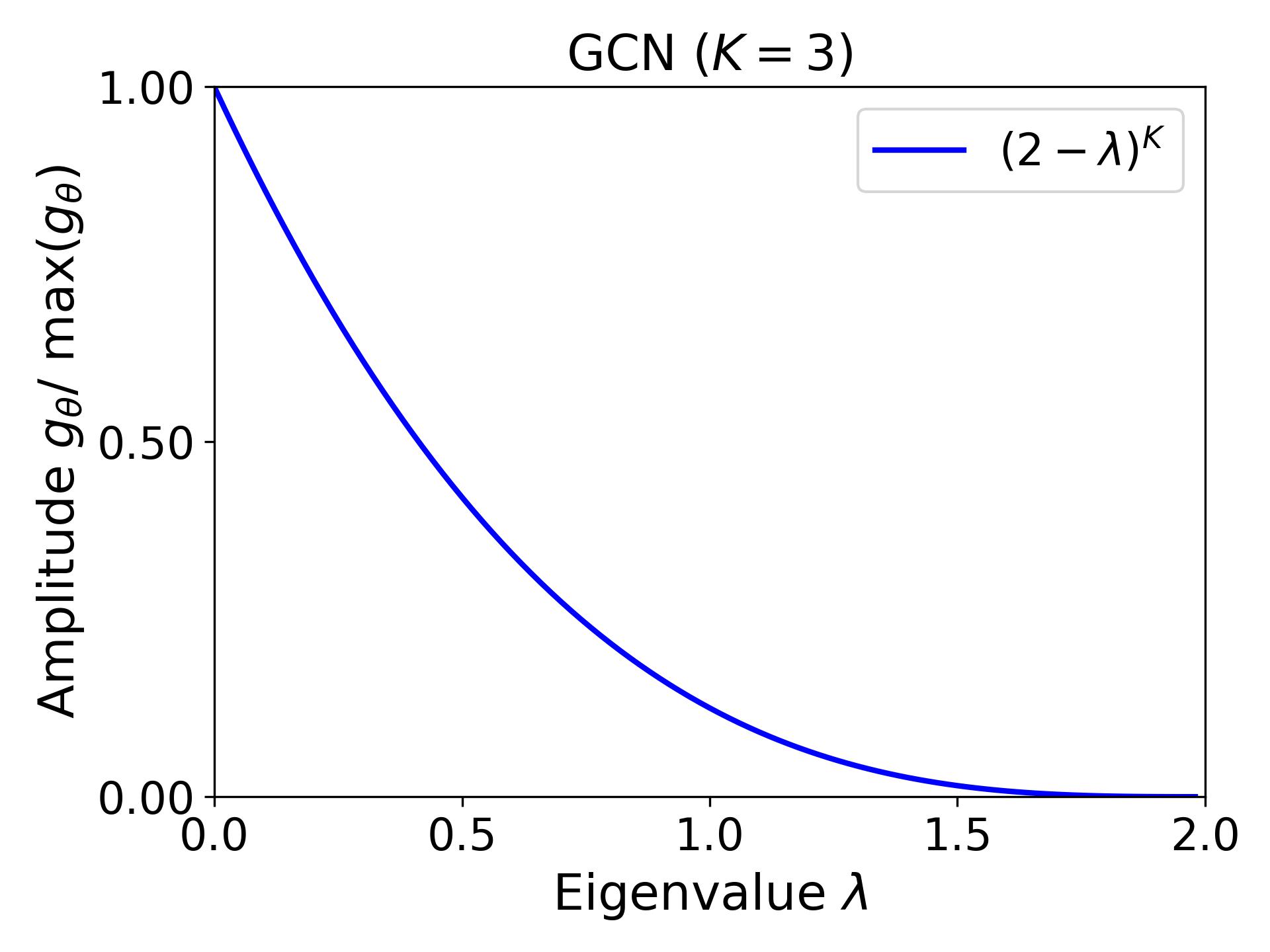}
	\end{minipage}
    }
    \subfloat[]{
	\begin{minipage}{0.475\linewidth}
		\centering
		\includegraphics[width=1\linewidth]{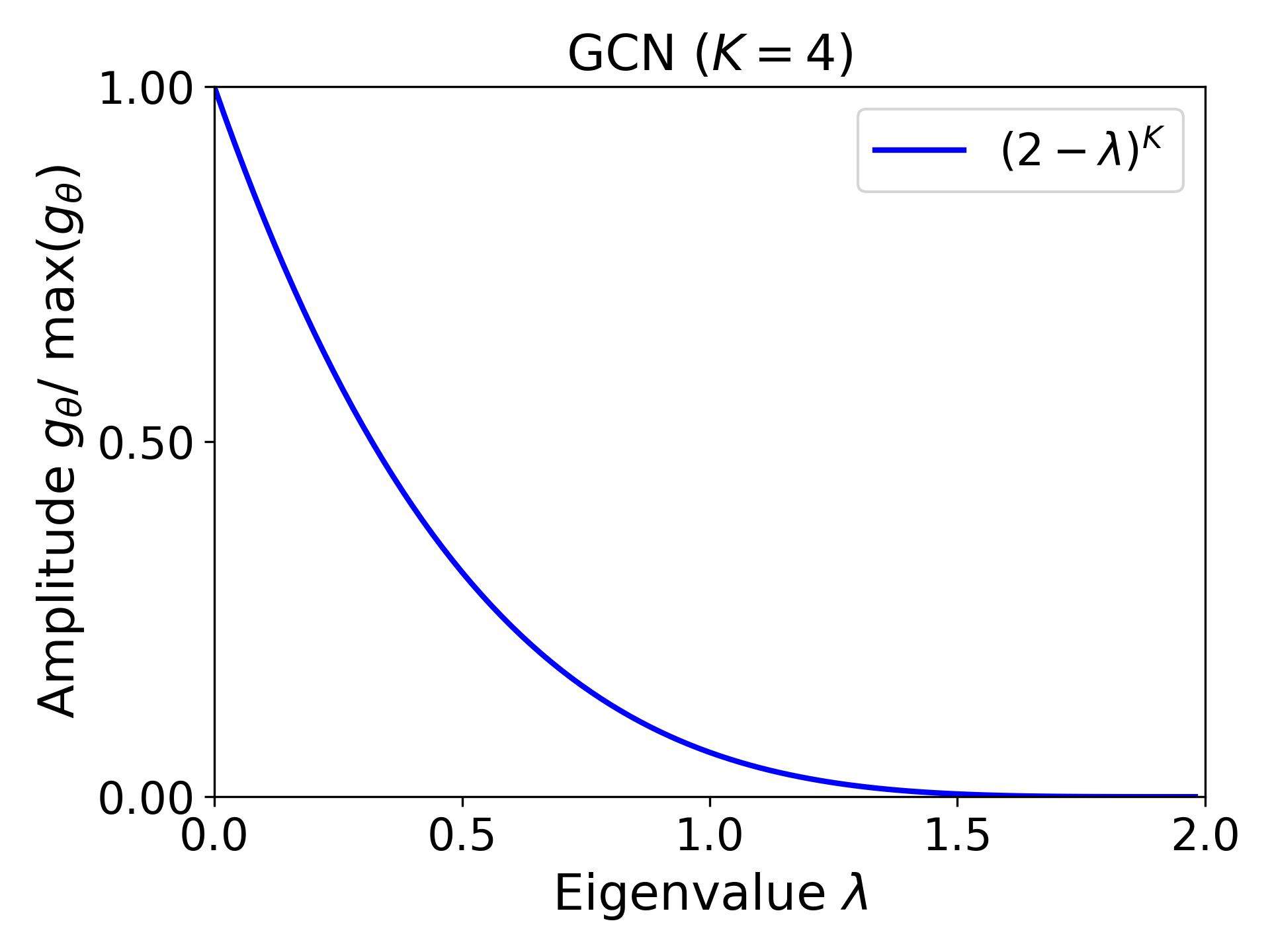}
	\end{minipage}
    }
\caption{Relations between eigenvalues and amplitudes in filter of GCN.}
\label{fig:3}
\end{figure}

According to Equ.~(\ref{equ:4}), the graph propagation of GCN can be formulated as follows:

\begin{equation}
\mathbf{H}_{GCN}=(\mathbf{2I}-\mathbf{L})^{K} \mathbf{X},
\end{equation}

\noindent where $K \in \mathbb{Z}^{+}$ denotes the number of graph convolution layers. The graph filter can be formulated as $g_{G C N}(\lambda)=(2-\lambda)^{K}$, $\lambda \in[0,2]$. 0 indicates low frequency information and 2 indicates high frequency information. The formula of GCN neighborhood polymerization is:

\begin{equation}
\tilde{\mathbf{h}}_{i}^{(l)}=\mathbf{h}_{i}^{(l)}+\sum_{j \in \mathcal{N}_{i}} \frac{1}{\sqrt{d_{i} d_{j}}} \mathbf{h}_{j}^{(l)}
\end{equation}

where $d_i$ and $d_j$ represent the degrees of nodes $v_i$ and $v_j$, respectively. The frequency responses of the first to fourth order GCN filters are shown in Fig.~\ref{fig:3} (a)-(d). GCN amplifies low-frequency signals and restrains high-frequency signals. Essentially, the GCN filter is a fixed low-pass filter with a greater tendency to aggregate low-frequency information. As the number of GCN layers increases, the order of the filter increases, and the suppression of high-frequency information is enhanced. Therefore, deep GCN models can lead to over-smoothing.

\subsection{Spectral Analysis for FAGCN}

\begin{figure}[htbp]
	\centering
    \subfloat[]{
	\begin{minipage}{0.475\linewidth}
		\centering
		\includegraphics[width=1\linewidth]{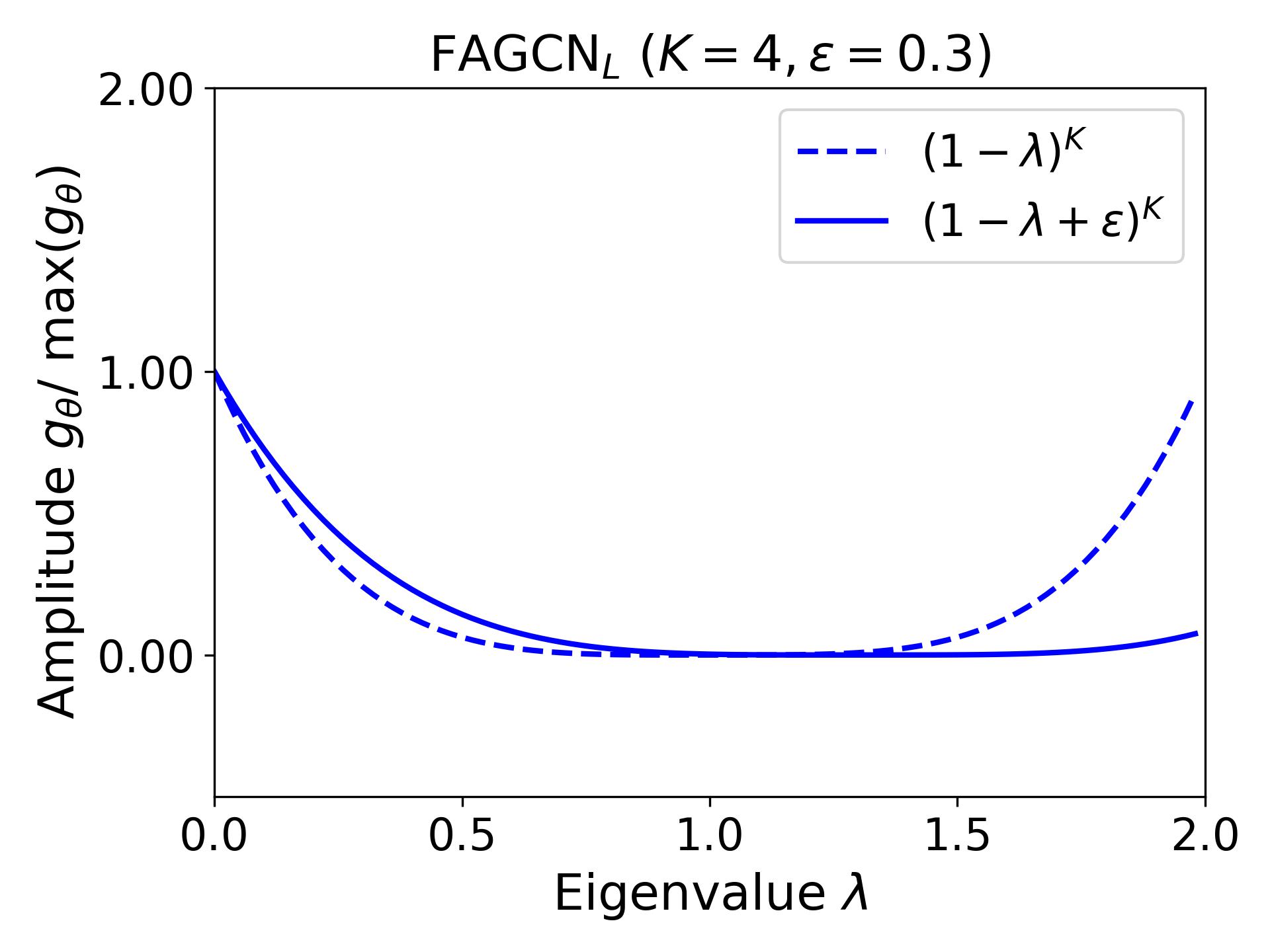}
	\end{minipage}
    }
    \subfloat[]{
	\begin{minipage}{0.475\linewidth}
		\centering
		\includegraphics[width=1\linewidth]{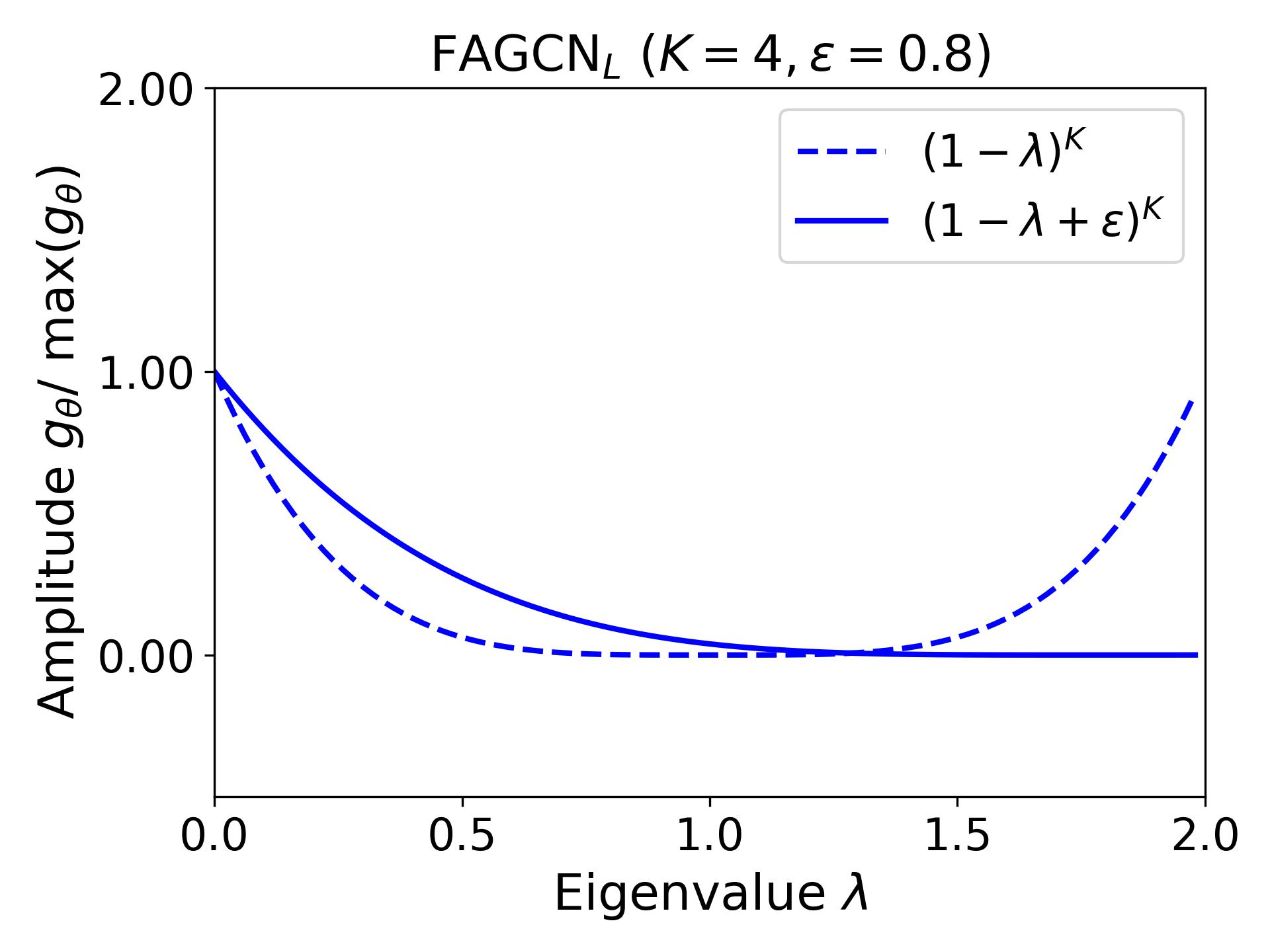}
	\end{minipage}
    }

	\subfloat[]{
	\begin{minipage}{0.475\linewidth}
		\centering
		\includegraphics[width=1\linewidth]{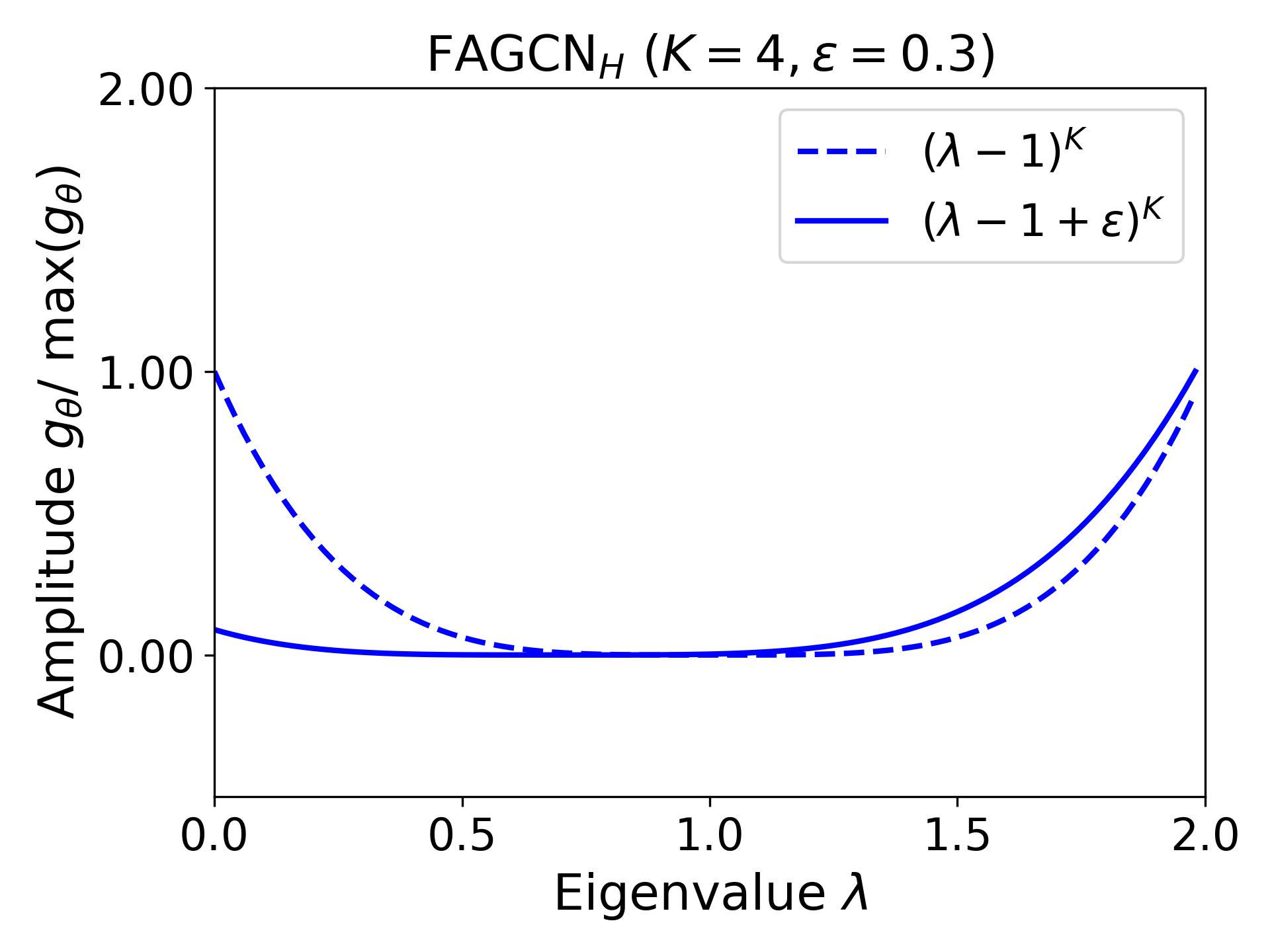}
	\end{minipage}
    }
    \subfloat[]{
	\begin{minipage}{0.475\linewidth}
		\centering
		\includegraphics[width=1\linewidth]{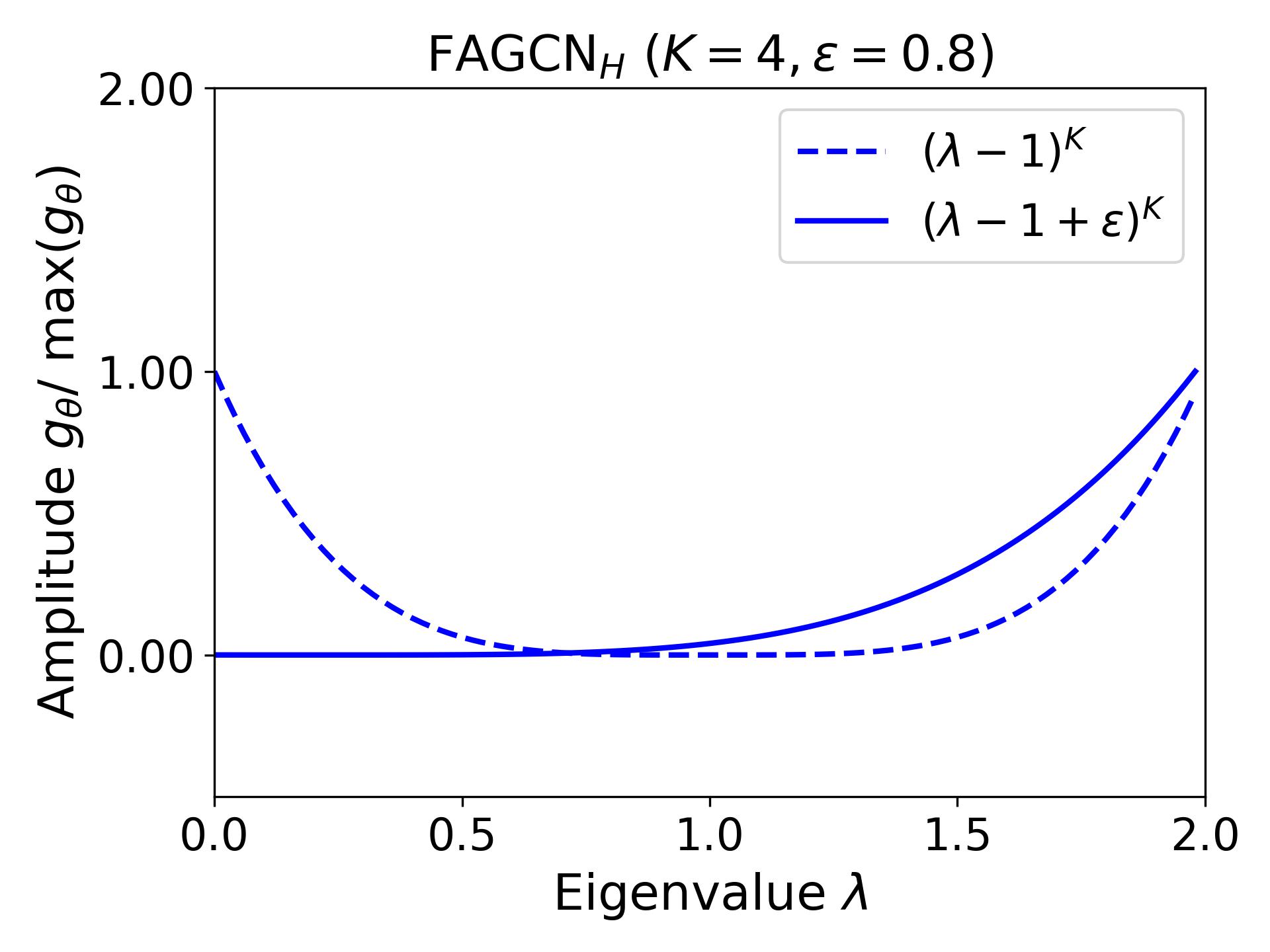}
	\end{minipage}
    }
\caption{Relations between eigenvalues and amplitudes in low-frequency and high-frequency filter of FAGCN.}
\label{fig:4}
\end{figure}

In order to extract low-frequency and high-frequency information separately, FAGCN incorporates two convolution kernels $\mathcal{F}_{L}$ and $\mathcal{F}_{H}$ to extract low-frequency and high-frequency information respectively:

\begin{equation}
\mathcal{F}_{L}=\varepsilon \mathbf{I}+\mathbf{D}^{-1 / 2} \mathbf{A} \mathbf{D}^{-1 / 2}=(\varepsilon+1) \mathbf{I}-\mathbf{L},
\end{equation}

\begin{equation}
\mathcal{F}_{H}=\varepsilon \mathbf{I}-\mathbf{D}^{-1 / 2} \mathbf{A} \mathbf{D}^{-1 / 2}=(\varepsilon-1) \mathbf{I}+\mathbf{L}.
\end{equation}

For a $K$-layer FAGCN model, its spectral filter is the combination of $g_{FAGCN_{L}}(\lambda)$ and $g_{FAGCN_H}(\lambda)$:

\begin{equation}
g_{FAGCN_{L}}(\lambda)=(1-\lambda+\epsilon)^{K},
\end{equation}

\begin{equation}
g_{FAGCN_H}(\lambda)=(\lambda-1+\epsilon)^{K},
\end{equation}

\noindent where $\epsilon \in[0,1]$. $g_{FAGCN_{L}}(\lambda)$ and $g_{FAGCN_{H}}(\lambda)$ denote low-frequency and high-frequency filters respectively. Fig.~\ref{fig:4} shows the frequency response of FAGCN$_L$ and FAGCN$_H$. FAGCN use the attention mechanism to learn the coefficients for low-frequency and high-frequency graph signals.

\begin{equation}
\begin{split}
\tilde{\mathbf{h}}_{i}^{(l)}&=\alpha_{i j}^{L}\left(\mathcal{F}_{L} \cdot \mathbf{H}^{(l)}\right)_{i}+\alpha_{i j}^{H}\left(\mathcal{F}_{H} \cdot \mathbf{H}^{(l)}\right)_{i}\\
&=\varepsilon \mathbf{h}_{i}^{(l)}+\sum_{j \in \mathcal{N}_{i}} \frac{\alpha_{i j}^{L}-\alpha_{i j}^{H}}{\sqrt{d_{i} d_{j}}} \mathbf{h}_{j}^{(l)}.
\end{split}
\end{equation}

Let $\alpha_{i j}^{G}=\alpha_{i j}^{L}-\alpha_{i j}^{H}$. The coefficient $\alpha_{i j}^{G}$ is normalized by the tanh function, which ranges from -1 to 1, FAGCN can adaptively learn low-frequency and high-frequency information. The filters in FAGCN are essentially linear combinations of $(1-\lambda+\epsilon)^{K}$ and $(\lambda-1+\epsilon)^{K}$. $\epsilon$ is actually a translation transformation of frequency response. Due to the limited range of values for $\epsilon$, the space for the filter to adjust is limited.

\subsection{Spectral Analysis for RFA-GNN}

\begin{figure}[htbp]
	\centering
    \subfloat[]{
	\begin{minipage}{0.475\linewidth}
		\centering
		\includegraphics[width=0.95\linewidth]{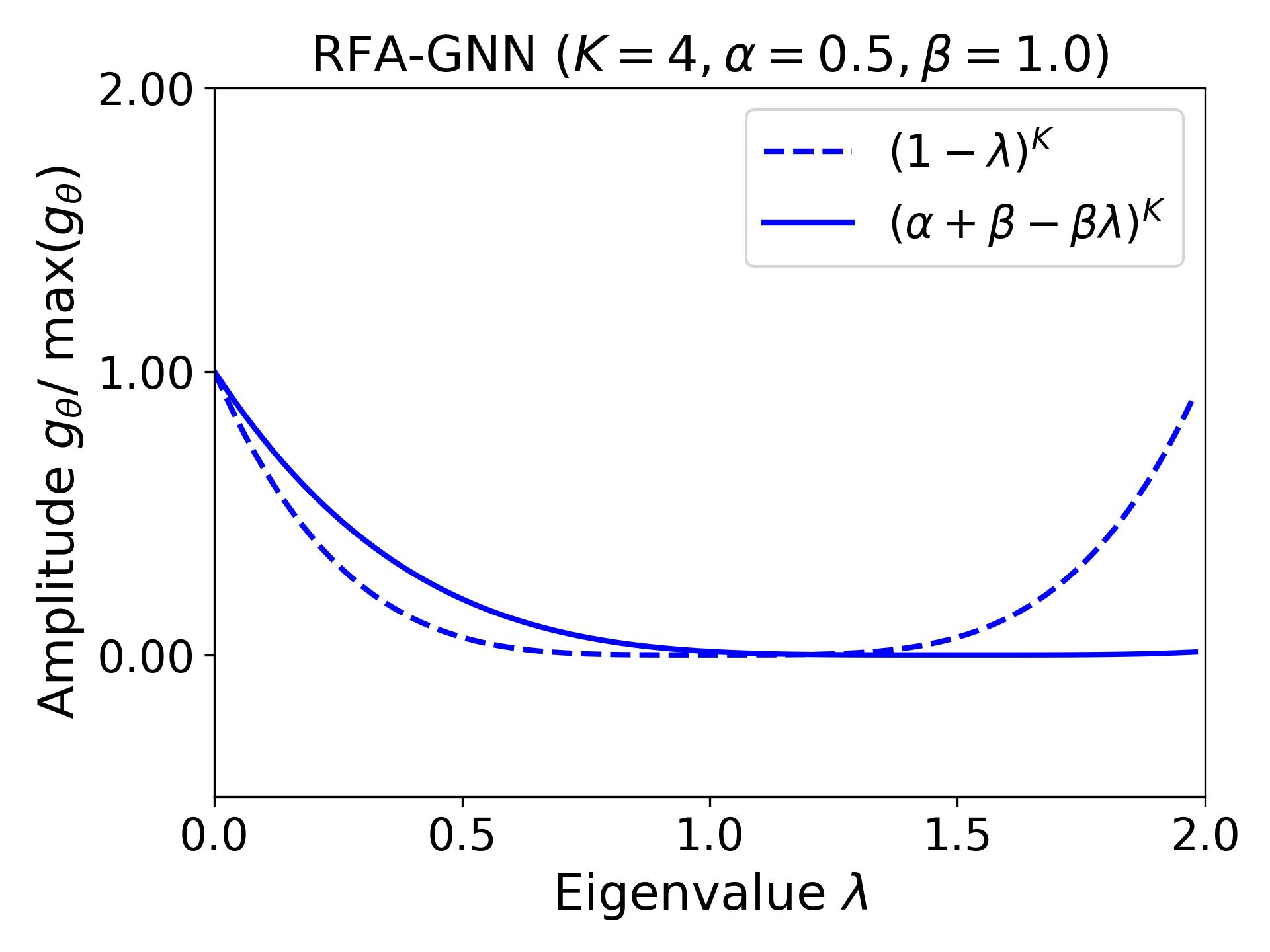}
	\end{minipage}
    }
    \subfloat[]{
	\begin{minipage}{0.475\linewidth}
		\centering
		\includegraphics[width=0.95\linewidth]{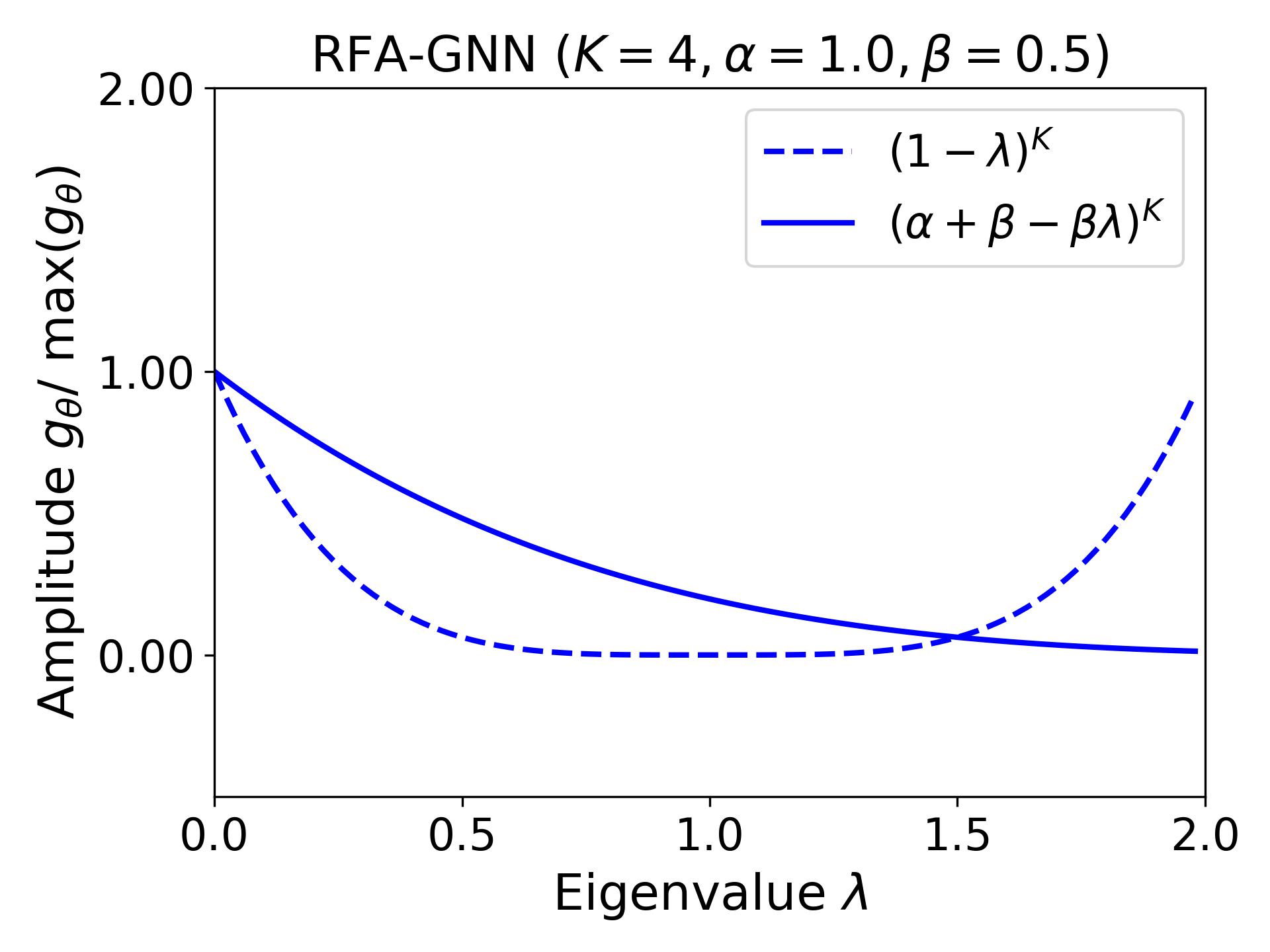}
	\end{minipage}
    }

	\subfloat[]{
	\begin{minipage}{0.475\linewidth}
		\centering
		\includegraphics[width=0.95\linewidth]{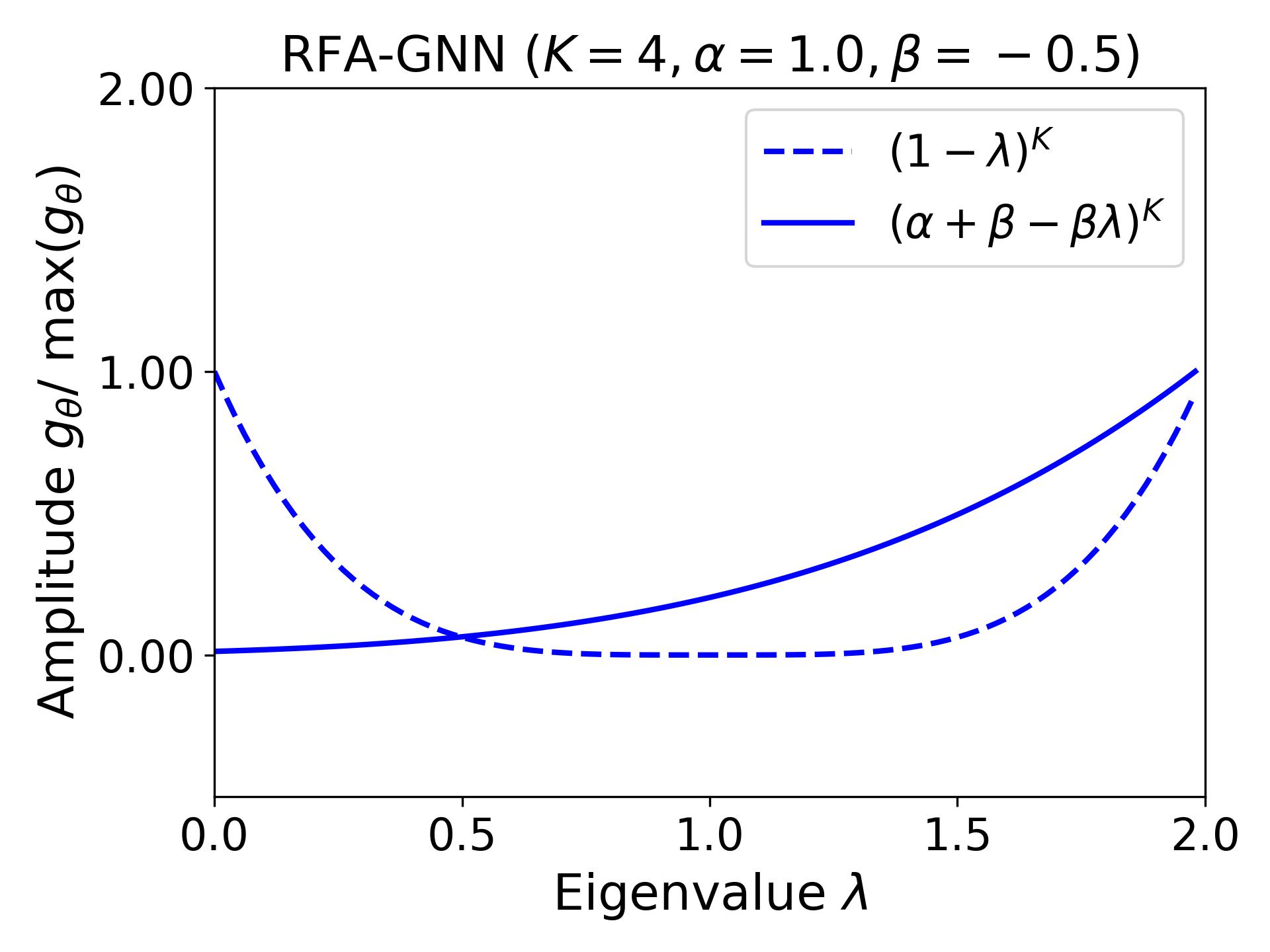}
	\end{minipage}
    }
    \subfloat[]{
	\begin{minipage}{0.475\linewidth}
		\centering
		\includegraphics[width=0.95\linewidth]{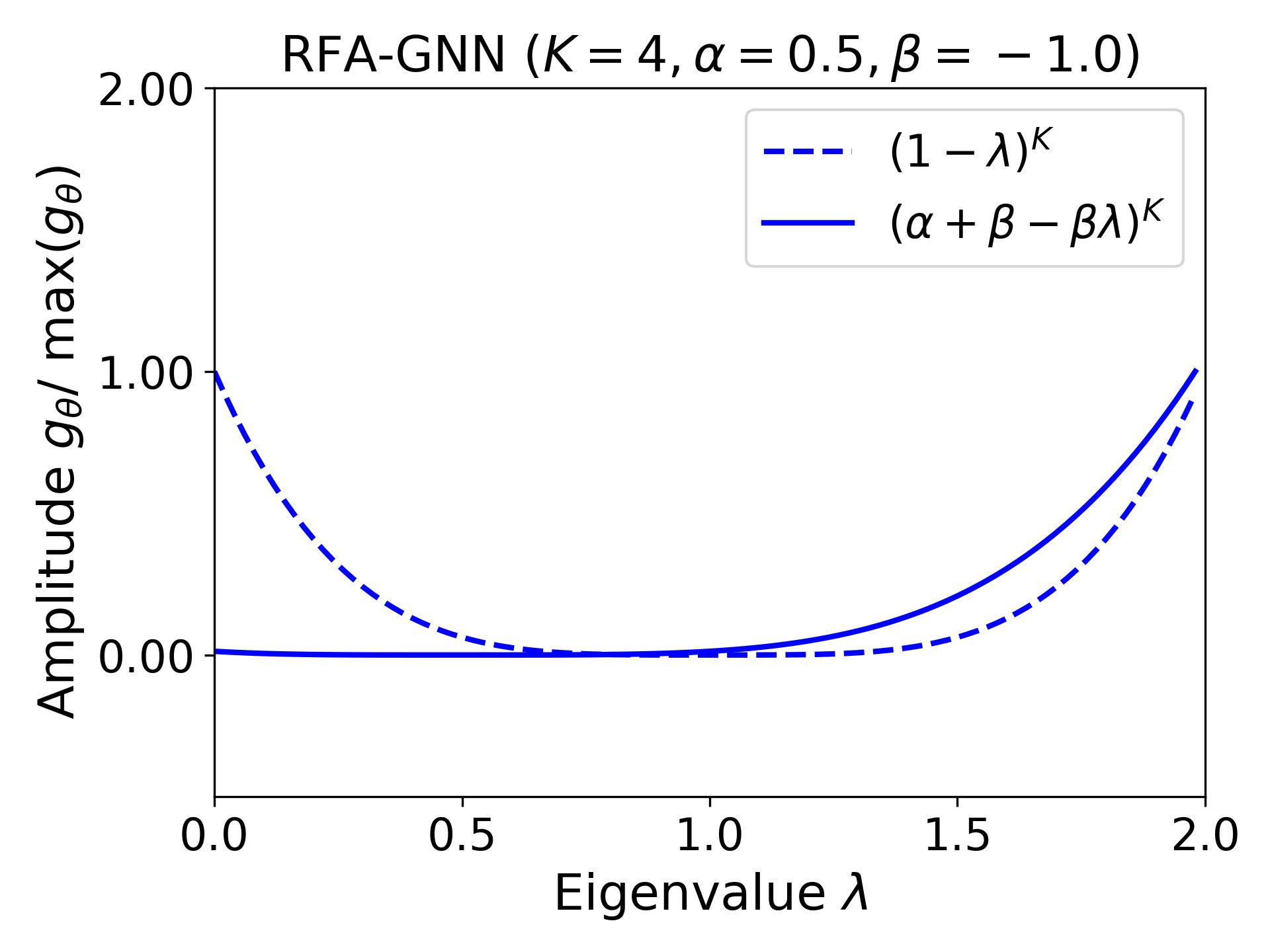}
	\end{minipage}
    }
\caption{Relations between eigenvalues and spectral amplitude for RFA-GNN.}
\label{fig:5}
\end{figure}

The RFA-GCN (frequency-adaptive graph convolutional network) is designed with a frequency-adaptive filter that includes a self-gating mechanism for adaptively selecting signals with different frequencies. RFA-GCN has a multi-hop relation-based frequency-adaptive architecture that considers both the graph properties of the data and high-order information between nodes. The convolution kernel of RFA-GNN is:

\begin{equation}
\mathcal{F}=\alpha \mathbf{I}+\beta \mathbf{D}^{-1 / 2} \mathbf{A D}^{-1 / 2}=(\alpha+\beta) \mathbf{I}-\beta \mathbf{L}.
\end{equation}

Its graph filter can be formulated as:

\begin{equation}
g_{RFA-GCN}(\lambda)=(\alpha+\beta-\beta \lambda)^{K},
\label{equ:RFAGCN}
\end{equation}

\noindent where $\alpha \in \left(0, 1\right]$ and $\beta\in \left(-1,  1\right)$. For the key parameter $\beta$ in Equ.~(\ref{equ:RFAGCN}), a shared adaptive mechanism was used to learn the frequency coefficient $\left\{\beta_{i, j}\right\}_{i, j=1}^{N}$ for each node. The formula of GCN neighborhood polymerization is:

\begin{equation}
\tilde{\mathbf{h}}_{i}^{(l)}=\alpha \mathbf{h}_{i}^{(l)}+\sum_{j \in \mathcal{N}_{i}} \frac{\beta_{i, j}^{(l)}}{\sqrt{d_{i} d_{j}}} \mathbf{h}_{j}^{(l)},
\end{equation}

\noindent and the frequency response of RFA-GCN of order $K$ can be written as:

\begin{equation}
(\alpha+\beta-\beta \lambda)^{K}=\beta^{K}\left(\frac{\alpha+\beta}{\beta}-\lambda\right)^{K}.
\end{equation}

The range of $\frac{\alpha+\beta}{\beta}$ is $(-\infty,+\infty)$.
Fig.~\ref{fig:5} shows the frequency response of RFA-GNN with different values of $\alpha$ and $\beta$. Although RFA-GNN extends RAGCN to more generalized cases, the frequency response of RFA-GNN is still a shifted transformation of $\left(-\lambda\right)^{K}$.

\subsection{Spectral Analysis for MSGS}

\begin{figure}[htbp]
	\centering
    \subfloat[]{
	\begin{minipage}{0.475\linewidth}
		\centering
		\includegraphics[width=0.95\linewidth]{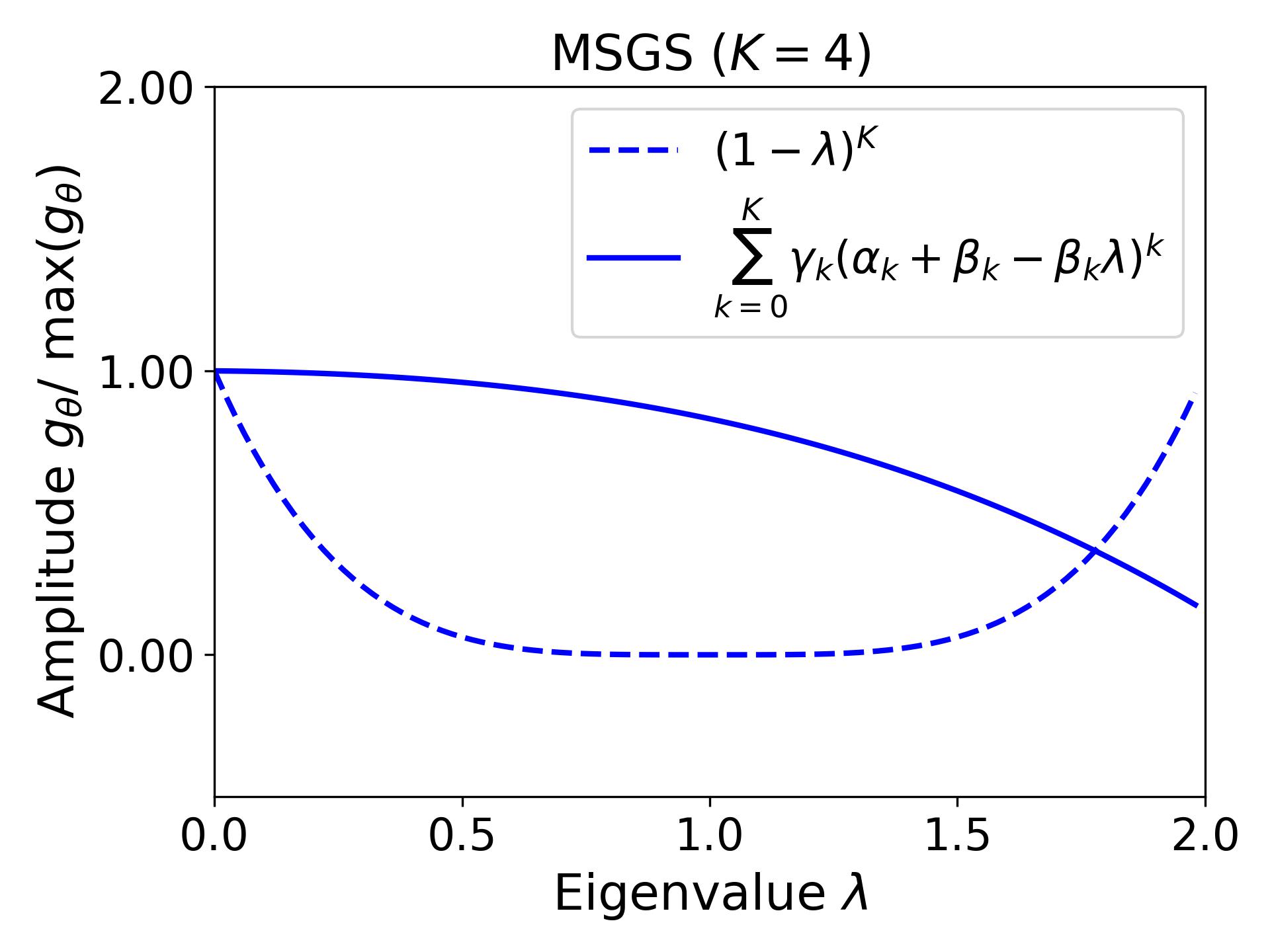}
	\end{minipage}
    }
    \subfloat[]{
	\begin{minipage}{0.475\linewidth}
		\centering
		\includegraphics[width=0.95\linewidth]{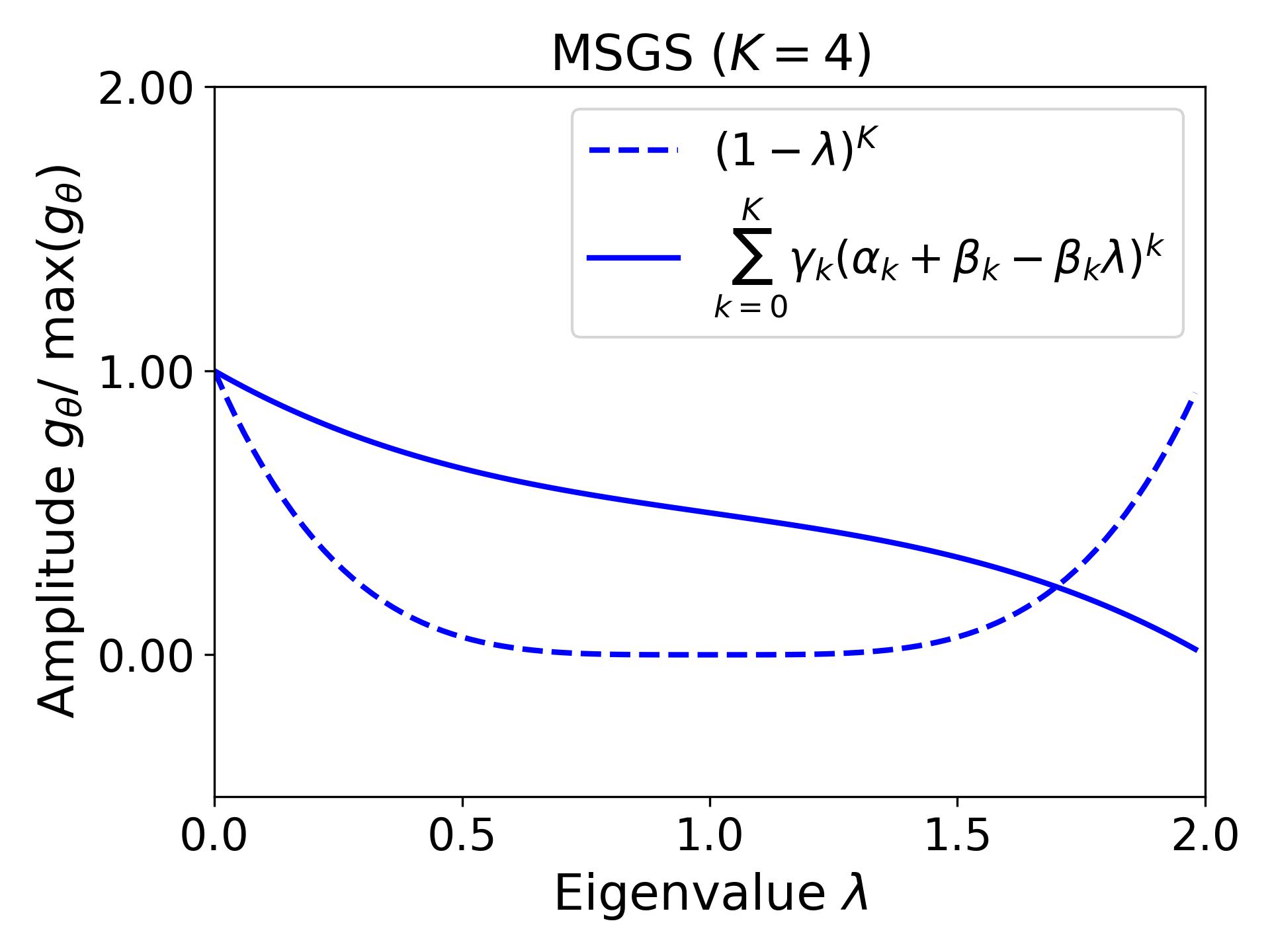}
	\end{minipage}
    }

	\subfloat[]{
	\begin{minipage}{0.475\linewidth}
		\centering
		\includegraphics[width=0.95\linewidth]{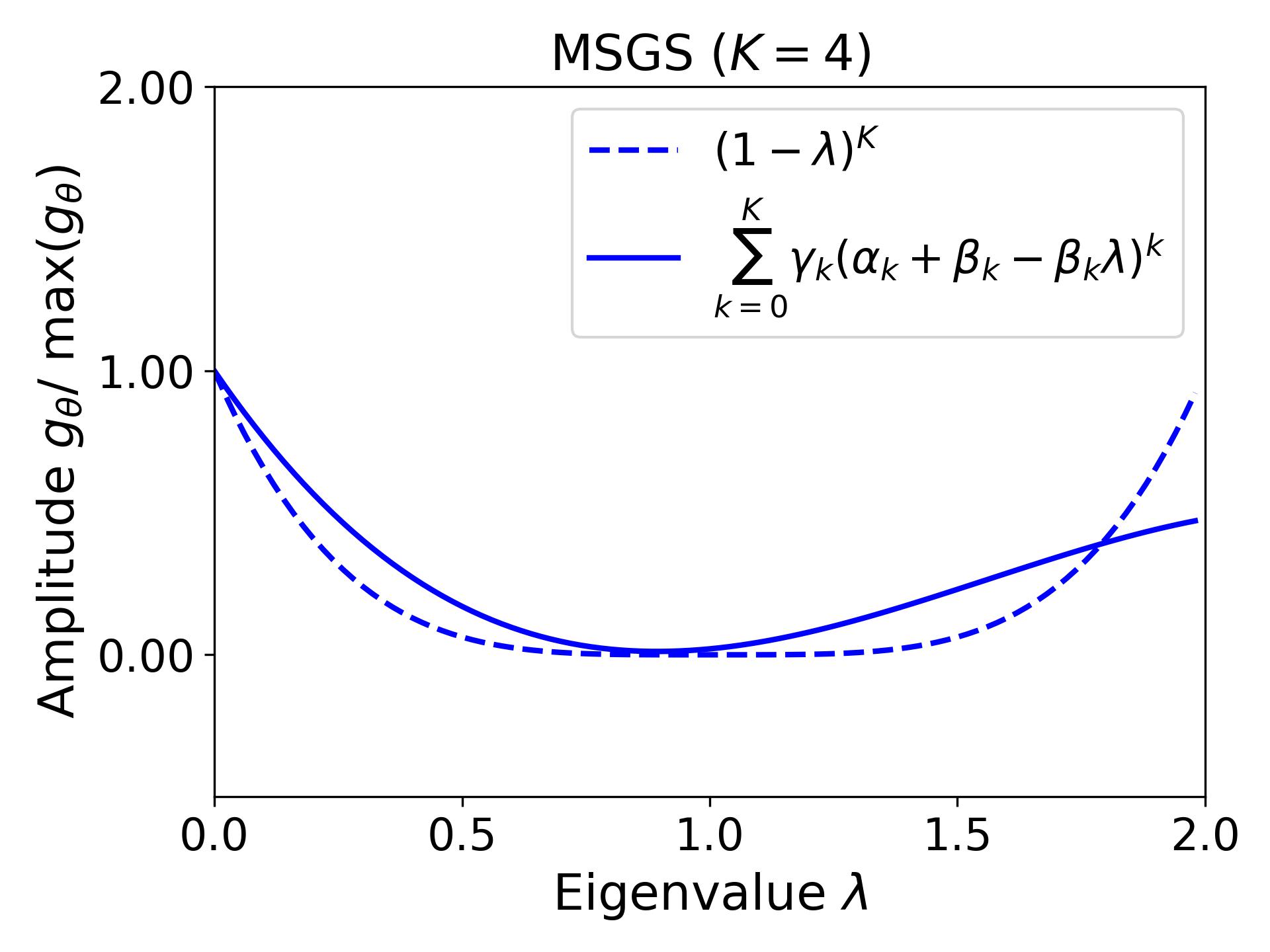}
	\end{minipage}
    }
    \subfloat[]{
	\begin{minipage}{0.475\linewidth}
		\centering
		\includegraphics[width=0.95\linewidth]{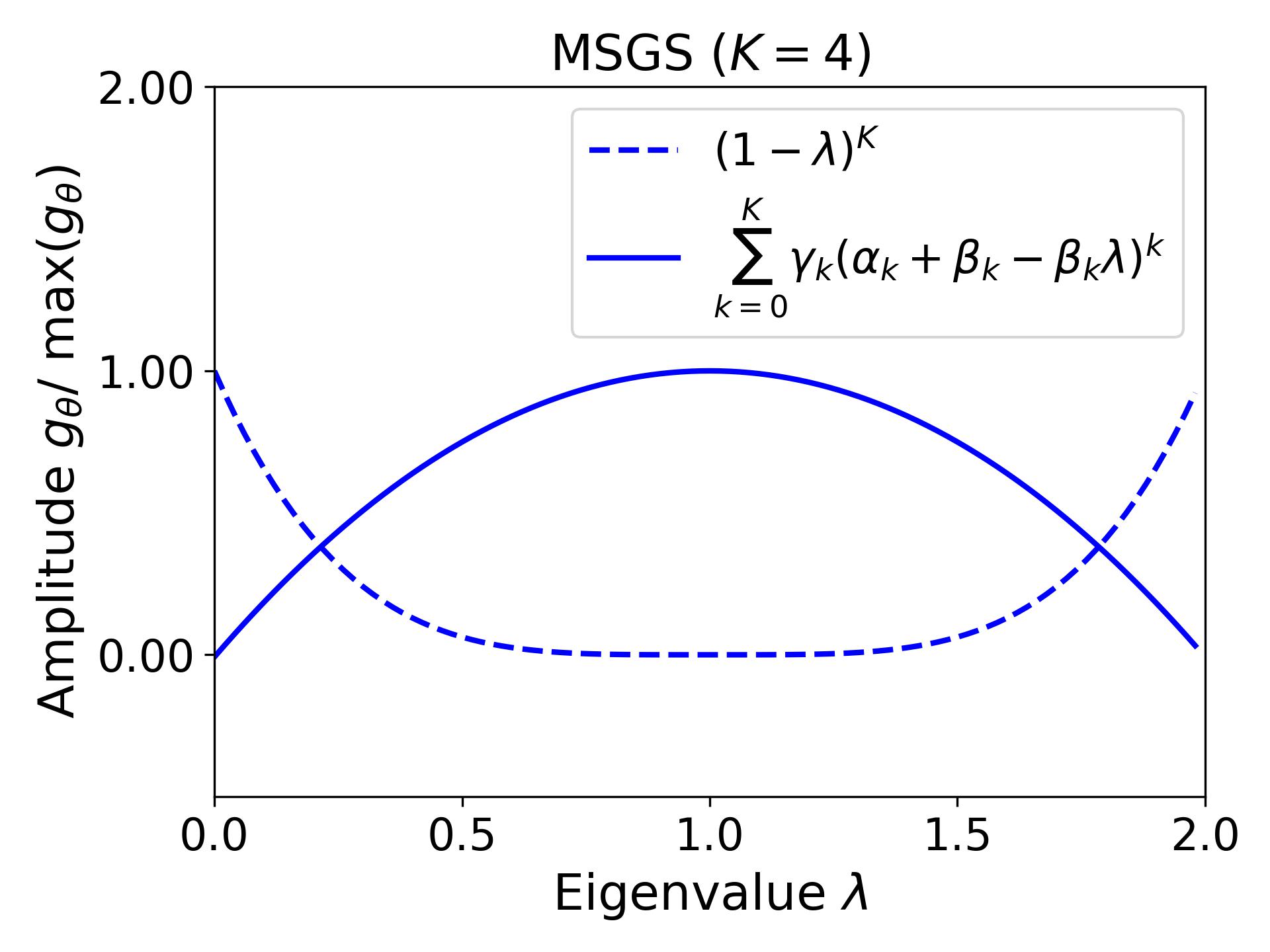}
	\end{minipage}
    }
\label{fig:6}
\caption{Relations between eigenvalues and spectral amplitude for MSGS.}
\end{figure}

The $K$-th order graph filter of MSGS can be formulated as follows:

\begin{equation}
\begin{split}
g_{MSGF}(\lambda)&=\sum_{k=0}^{K} \gamma_{k}(\alpha^{(k)}+\beta^{(k)}-\beta^{(k)} \lambda)^{k}\\
&=\sum_{k=0}^{K}\gamma_{k}(\beta^{(k)})^{k}(\frac{\alpha^{(k)}+\beta^{(k)}}{\beta^{(k)}}-\lambda)^{k},
\end{split}
\end{equation}

\noindent where $\alpha_{k} \in(0,1]$, $\beta_{k} \in(-1,1)$. The parameters  and  of MSGS can be adjusted to utilize different frequencies from the $K$-hop neighborhood.

\begin{equation}
\mathbf{p}_{i}=\sum_{k=0}^{K} \gamma_{k, i} \mathbf{z}_{i}^{(k)}=\sum_{k=0}^{K} \gamma_{k, i}\left[\alpha_{i}^{(k)} \mathbf{h}_{i}^{(k)}+\sum_{j \in \mathcal{N}_{i}} \frac{\beta_{i, j}^{(k)}}{\sqrt{d_{i} d_{j}}} \mathbf{h}_{j}^{(k)}\right].
\end{equation}

\begin{figure}[htbp]
	\centering
	\includegraphics[width=0.9\linewidth]{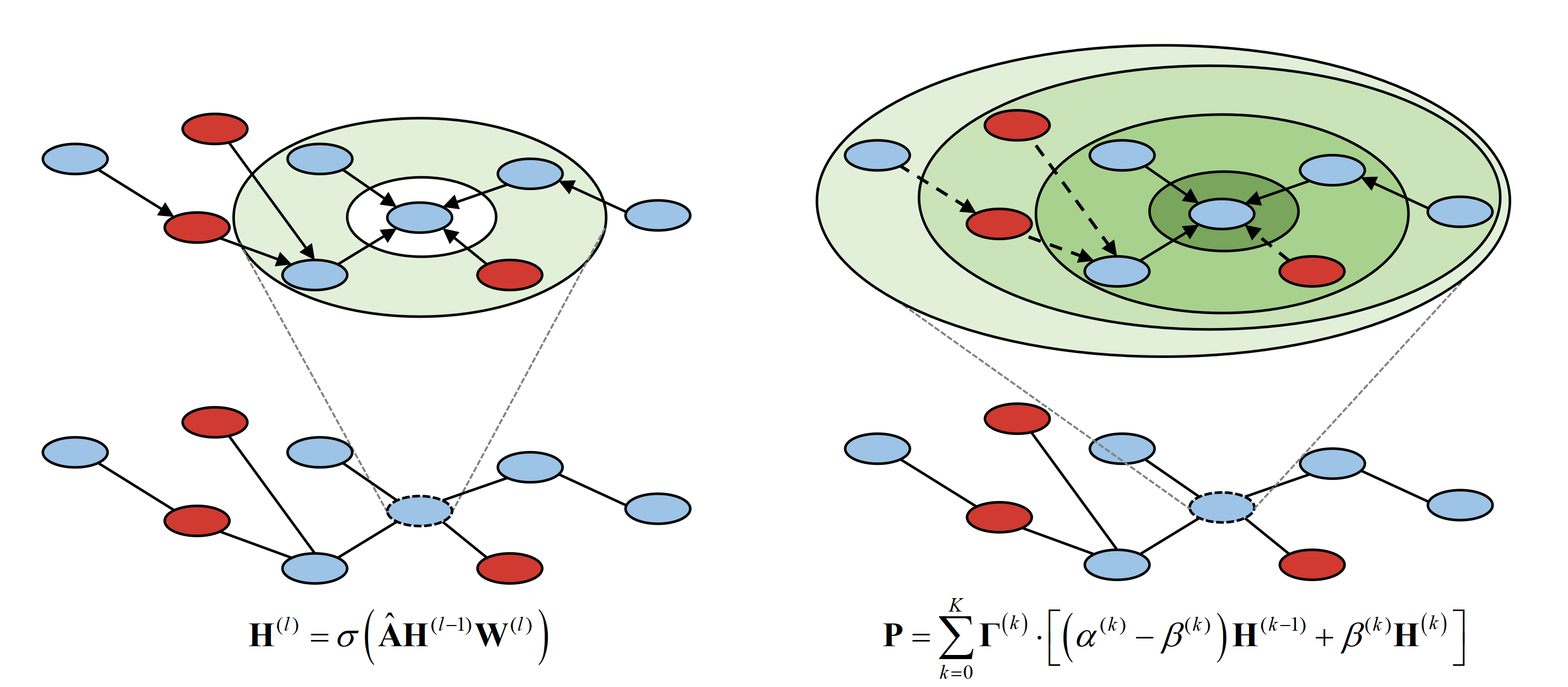}
	\caption{RFA-GNN2}
	\label{fig:MGSG_aggregate}
\end{figure}

As shown in Fig.~\ref{fig:MGSG_aggregate}, the frequency response of $K$-layer GCN, RAGCN, and RFA-GNN only considers the $K$-th power of $\lambda$, and the frequency response of graph filters is relatively fixed. Compared to the aforementioned methods, MSGS expands the frequency response to a $K$-order polynomial, allowing for more flexible adaptation of low and high-frequency information.

\noindent \textbf{Proposition 3} For a single MSGS graph filter $g$, $C*g$ can represent any $K$-th order polynomial, where $C$ is any real number.
This proposition highlights that the frequency response of $K$-layer MSGS can represent any $K$-th order polynomial, which expands the space of graph filters. As a result, the model can be more flexible in preserving or filtering out low-frequency and high-frequency information.

\begin{equation}
C*g_{MSGF}(\lambda)=\sum_{k=0}^{K} C \gamma_{k}(\beta^{(k)})^{k}(\frac{\alpha^{(k)}+\beta^{(k)}}{\beta^{(k)}}-\lambda)^{k}
\end{equation}

Let $c_{1}=C \gamma_{k}(-\beta^{(k)})^{k}$, $c_{2}=\frac{\alpha^{(k)}+\beta^{(k)}}{\beta^{(k)}}$

\begin{equation}
\begin{split}
C*g_{MSGF}(\lambda)&=\sum_{k=0}^{K} C \gamma_{k}(-\beta^{(k)})^{k}(\lambda-\frac{\alpha^{(k)}+\beta^{(k)}}{\beta^{(k)}})^{k}\\
&=\sum_{k=0}^{K} c_{1}(\lambda-c_{2})^{k},
\end{split}
\end{equation}

\noindent where $c_{1}, c_{2} \in(-\infty,+\infty)$, therefore, $C*g$ can represent any $K$-th order polynomial expression.
The frequency response of MSGS under different parameters is shown in Fig.~\ref{fig:6}. Compared with the previous GNNS, MSGS has a larger variation space and can learn a more accurate frequency response. MSGS can adaptively utilize the information of the $K$-hop neighborhood of the target node. By learning the weights of the edges during adaptive neighborhood aggregation, positive weights are assigned to edges with low-frequency information to enhance the information through addition. In contrast, negative weights are assigned to those with high-frequency information for enhancement through subtraction. This approach strengthens the low-frequency information and enhances the high-frequency information in the graph.

\section{Experiment setup}
\subsection{Dataset}
We evaluated MSGS and other bot detection models on three datasets: Cresci-15~\cite{article03}, Twibot-20~\cite{article35}, and MGTAB~\cite{article12}. These datasets provide information on the follower and friend relationships between users. Cresci-15 is a dataset of 5,301 users labeled genuine or automated accounts. Twibot-20 is a dataset of 229,580 users and 227,979 edges, of which 11,826 accounts have been labeled genuine or automated. MGTAB is a dataset containing more than 1.5 million users and 130 million tweets. It provides information on seven types of relationships between these users and labels 10,199 accounts as either genuine or bots.
We constructed user social graphs by using all labeled users and follower and friend relationships between them. For MGTAB, we used the top 20 user attribute features with the highest information gain and 768-dimensional user tweet features extracted by BERT as user features. For Twibot-20, following~\cite{article15}, we used 16 user attribute features, user description features, and user tweet features extracted by BERT. For Cresci-15, as described in~\cite{article16}, we used 6 user attribute features, 768-dimensional user description features extracted by BERT, and user tweet features. Table~\ref{tb:Statistics} provides a summary of the dataset statistics. We randomly partitioned all datasets using a 1:1:8 ratio.

\begin{table}[h]
\caption{Statistics of datasets used in the paper.}
\begin{center}
\begin{adjustbox}{width=0.90\linewidth}
\begin{tabular}{|l|c|c|c|c|c|}
\hline
Dataset & Nodes	&Bots	&Human	&Edges	&Features \\
\hline
Cresci-15  &5,301	&3,351	&1,950	&14,220	 &1,542 \\
\hline
Twibot-20  &11,826	&6,589	&5,237	&15,434	 &1,553 \\
\hline
MGTAB      &10,199	&2,748	&7,451	&1,700,108	&788 \\
\hline
\end{tabular}
\end{adjustbox}
\end{center}
\label{tb:Statistics}
\end{table}

\subsection{Baseline Methods}
To verify the effectiveness of our proposed RF-GNN, we compare it with various semi-supervised learning baselines. The detail about these baselines as described as follows:
\begin{itemize}
\item \textbf{Node2Vec}~\cite{article27} is a weighted random walk algorithm that facilitates the creation of node vectors that satisfy both homophily and structural similarity assumptions.
\item \textbf{APPNP}~\cite{article14} combines GCN with PageRank to better propagate information from neighboring nodes, utilizing a large, adjustable neighborhood.
\item \textbf{GCN}~\cite{article17} is a spectral graph convolution method that generates node embedding vectors by truncating the Chebyshev polynomial to the first-order neighborhoods.
\item \textbf{SGC}~\cite{article18} is a simplified version of GCN that reduces excessive complexity by iteratively removing non-linearities between GCN layers and collapsing the resulting function into a single linear transformation.
\item \textbf{GAT}~\cite{article19} is a semi-supervised homogeneous graph model that employs the attention mechanism to determine the weights of node neighborhoods, thereby improving the performance of graph neural networks.
\item \textbf{Boosting-GNN}~\cite{article20} trains a series of GNN base classifiers by serializing them, and sets higher weights for training samples that are not correctly classified by previous classifiers, thus obtaining higher classification accuracy and better reliability.
\item \textbf{LA-GCN}~\cite{article23} improves the expressiveness of GNN by learning the conditional distribution of neighbor features to generate features.
\item \textbf{JK-Nets}~\cite{article21} is a kind of GNN that employs jump knowledge to obtain a more effective structure-aware representation by flexibly utilizing the distinct neighborhood ranges of each node.
\item \textbf{MSGCN}~\cite{article08} adds multi-scale information to the neural network and fuses it with a self-attention mechanism and multi-scale information into the GCN design. This enhances the neural network's expression ability and alleviates the over-smoothing phenomenon of GCNs.
\item \textbf{FAGCN}~\cite{article04} explored, for the first time, the role of low-frequency and high-frequency signals in GNNs. They then designed a novel frequency-adaptive GCN that combines low-frequency and high-frequency signals in an adaptive manner.
\item \textbf{RFA-GNN}~\cite{article05} designs a frequency-adaptive filter with a self-gating mechanism that picks signals with different frequencies adaptively, without knowing the heterophily levels.
\item \textbf{AdaGNN}~\cite{article06} is an adaptive frequency response filter that can learn to control information flow for different feature channels. It adjusts the importance of different frequency components for each input feature channel, which creates a learnable filter when multiple layers are stacked together.
\end{itemize}

\subsection{Parameter Settings and Hardware Configuration}
All baseline methods have been initialized using the recommended parameters from their official codes and have undergone meticulous fine-tuning. Additionally, we conducted training for 500 epochs and selected the model with the highest validation accuracy for testing. Our model was trained using the Adam optimizer for 500 epochs. We experimented with different learning rates, specifically $\{$0.001, 0.005, 0.01$\}$. The number of layers, K, was set to 10 for all datasets. The L2 weight decay factor of 5e-4 was applied across all datasets. The dropout rate ranged from 0 to 0.5. The model presented in this paper utilized hidden units of $\{$16, 32, 64, 128$\}$. We fine-tuned the remaining parameters until achieving optimal classification performance.

We implemented MSGS using PyTorch 1.8.0 and Python 3.7.10, along with PyTorch Geometric~\cite{article34} for efficient sparse matrix multiplication. All experiments were executed on a server equipped with 9 Titan RTX GPUs, an Intel Xeon Silver 4210 CPU running at 2.20GHz, and 512GB of RAM. The operating system employed was Linux bcm 3.10.0.

\subsection{Evaluation Metrics}
We employ both accuracy and F1-score to assess the overall performance of the classifier.

\begin{equation}
\label{eq:Accuracy}
\text{Acurracy}=\frac{TP+TN}{TP+FP+FN+TN},
\end{equation}

\begin{equation}
\label{eq:Precision}
\text{Precision}=\frac{TP}{TP+FP},
\end{equation}

\begin{equation}
\label{eq:Recall}
\text{Recall}=\frac{TP}{TP+FN},
\end{equation}

\begin{equation}
\label{eq:F1}
\text{F1}=\frac{2 \times \text{Precision} \times \text {Recall}}{\text {Precision}+ \text {Recall}},
\end{equation}

\noindent where $TP$ is True Positive, $TN$ is True Negative, $FP$ is False Positive, $FN$ is False Negative.

\section{Experiment results}
In this section, we performance experiments on real world social bot detection benchmarks to evaluate MSGS. We aim to answer the following questions:
\begin{itemize}
\item \textbf{Q1:} How does MSGS perform compare to the state-of-the-art baselines in different scenarios? (Section~\ref{sec:Evaluation_real}).
\item \textbf{Q2:} How does MSGS perform under different training set partitions? (Section~\ref{sec:Evaluation_real}).
\item \textbf{Q3:} How does each individual module contributes to the performance of MSGS? (Section~\ref{sec:Ablation}).
\item \textbf{Q4:} Can MSGS alleviate the over-fitting phenomenon prevalent in GNNs? (Section~\ref{sec:Over-smoothing}).
\item \textbf{Q5:} Can MSGS effectively use high and low-frequency information? What are the differences in using high-frequency and low-frequency information across different datasets? (Section~\ref{sec:Coefficients}).
\item \textbf{Q6:} What are the frequency responses learned by MSGS on different datasets? (Section~\ref{sec:Filter}).
\end{itemize}

\subsection{Evaluation on the Real-World Dataset}
\label{sec:Evaluation_real}
In this section, we perform experimental analysis on publicly available social bot detection datasets, aimed at assessing the efficacy of our proposed method. The data was partitioned randomly into training, validation, and test sets, maintaining a ratio of 1:1:8. To ensure reliability and minimize the impact of randomness, we performed five evaluations of each method using different seeds. Our results are reported in Table~\ref{tb:main_results}, illustrating the average performance of the baselines, as well as our proposed method, MSGS, and its various adaptations. Notably, MSGS consistently outperforms both the baselines and the alternative variants across all scenarios.

\begin{table*}[t]
\caption{Comparison of the average performance of different methods for social bot detection. The best result of the baseline method and the complete MGSG proposed by us is highlighted in bold.}
\begin{center}
\setlength{\columnsep}{1pt}%
\begin{adjustbox}{width=0.925\linewidth}
\begin{tabular}{@{\extracolsep{1pt}}|r|l|c|c|c|c|c|c|@{}}
\hline
& \multirow{2}{*}{\textbf{Method}} & \multicolumn{2}{c|}{MGTAB} & \multicolumn{2}{c|}{Twibot-20} & \multicolumn{2}{c|}{Cresci-15}  \\
\cline{3-8}
\rule{0pt}{2.2ex}
& & Acc & F1 & Acc & F1 & Acc & F1 \\
\cline{2-8}
\rule{0pt}{2.5ex}
\multirow{13}{*}{\rotatebox{90}{Baseline}}
& Node2Vec         & 73.35$\pm$0.19 & 60.20$\pm$0.72 & 51.85$\pm$0.20
                   & 48.98$\pm$0.39 & 73.22$\pm$0.60 & 70.83$\pm$0.56
             \\
\cline{2-8}
& APPNP            & 75.08$\pm$1.73 & 61.66$\pm$1.25 & 53.13$\pm$3.80
                   & 50.82$\pm$3.38 & 95.33$\pm$0.48 & 94.97$\pm$0.51
             \\
\cline{2-8}
& GCN              & 84.98$\pm$0.70 & 79.63$\pm$1.02 & 67.76$\pm$1.24
                   & 67.34$\pm$1.16 & 95.19$\pm$0.99 & 94.88$\pm$1.02
             \\
\cline{2-8}
& SGC              & 85.14$\pm$0.72 & 80.60$\pm$1.66 & 68.01$\pm$0.40
                   & 67.60$\pm$0.24 & 95.69$\pm$0.84 & 95.39$\pm$0.85
             \\
\cline{2-8}
& GAT              & 84.94$\pm$0.29 & 80.22$\pm$0.44 & 71.71$\pm$1.36
                   & 71.18$\pm$1.38 & 96.10$\pm$0.46 & 95.79$\pm$0.49
             \\
\cline{2-8}
& Boosting-GNN     & 85.14$\pm$0.72 & 79.84$\pm$1.09 & 68.10$\pm$0.77
                   & 67.77$\pm$0.79 & 95.69$\pm$0.47 & 95.40$\pm$0.49
             \\
\cline{2-8}
& LA-GCN           & 85.50$\pm$0.28 & 81.12$\pm$0.42 & 74.36$\pm$0.67
                   & 73.49$\pm$0.67 & 96.02$\pm$0.39 & 95.70$\pm$0.43
             \\
\cline{2-8}
& Mixhop	       & 85.70$\pm$1.09	& 81.79$\pm$1.01 & 77.56$\pm$1.88	
                   & 77.22$\pm$1.84	& \textbf{96.40$\pm$0.28} & \textbf{96.11$\pm$0.31}
            \\
\cline{2-8}
& JK-Nets          & 84.58$\pm$0.28 & 80.60$\pm$0.78 & 71.01$\pm$0.54
                   & 70.77$\pm$0.37 & 96.04$\pm$0.42 & 95.76$\pm$0.43
             \\
\cline{2-8}
& MSGCN	           & 86.17$\pm$0.51	& 83.00$\pm$0.96 & 75.79$\pm$1.67	
                   & 75.21$\pm$1.94	& 96.10$\pm$0.48 & 95.78$\pm$0.51
            \\
\cline{2-8}
& FAGCN	           & 86.02$\pm$0.01	& 81.92$\pm$0.90 & 78.45$\pm$0.29
	               & 77.95$\pm$0.29	& 95.82$\pm$0.21 & 95.63$\pm$0.32
            \\
\cline{2-8}
& RFA-GNN	       & \textbf{86.20$\pm$0.34}	& \textbf{83.03$\pm$0.31} & \textbf{80.39$\pm$0.28}	
                   & \textbf{80.12$\pm$0.24}	& 96.12$\pm$0.32 & 95.83$\pm$0.28
            \\
\cline{2-8}
& AdaGNN	       & 86.14$\pm$0.52	& 81.84$\pm$0.94 & 78.10$\pm$0.53	
                   & 77.12$\pm$0.69 & 95.78$\pm$0.43 & 95.82$\pm$0.32
            \\
\hline
\hline
\multirow{4}{*}{\rotatebox{90}{Ablation}}

&MSGS w/o SAM(N)   &86.42$\pm$0.33	&82.26$\pm$0.62	&80.95$\pm$1.04	
                   &80.58$\pm$0.86	&96.05$\pm$0.52	&95.74$\pm$0.38
                    \\
\cline{2-8}
&MSGS w/o SAM(S)	&87.16$\pm$0.28	&83.37$\pm$0.47	&81.16$\pm$0.62	
                    &80.85$\pm$0.65	&96.20$\pm$0.36	&96.07$\pm$0.39
                    \\
\cline{2-8}
&MSGS w/o MS	    &86.84$\pm$0.31	&82.68$\pm$1.15	&80.49$\pm$0.32	
                    &80.29$\pm$0.35	&96.08$\pm$0.25	&95.89$\pm$0.34
                    \\
\cline{2-8}
&MSGS	            &\textbf{87.71$\pm$0.31}	&\textbf{84.14$\pm$0.68}	&\textbf{82.33$\pm$0.65}	
                    &\textbf{81.92$\pm$0.51}	&\textbf{96.59$\pm$0.24}	&\textbf{96.27$\pm$0.27}
                    \\
\hline
\end{tabular}
\end{adjustbox}
\end{center}
\label{tb:main_results}
\end{table*}

MSGS demonstrates significantly superior performance compared to GCN across all datasets. Specifically, MSGS exhibits improvements of 4.13\%, 14.57\%, and 1.40\% on the MGTAB, Twibot-20, and Cresci-15 datasets, respectively, when compared to the baseline model GCN. Notably, detecting bots on the Cresci-15 dataset proves to be relatively facile, as most detection methods achieve over 95\% accuracy. Consequently, there is limited scope for enhancement on this dataset. Furthermore, in comparison to the best results among state-of-the-art methods, our approach enhances accuracy by 1.51\%, 1.94\%, and 0.19\% on the MGTAB, Twibot-20, and Cresci-15 datasets, respectively. These outcomes effectively demonstrate the efficacy of MSGS.

Regarding the multi-scale GNN, JK-Net incorporates skip connections between different layers, enabling the collection and aggregation of feature representations from diverse hierarchical levels to form the final feature representation. This approach retains more information compared to GCN. MSGCN, on the other hand, leverages information from multi-order neighborhoods, leading to respective improvements of 2.59\%, 8.03\%, and 0.91\% on the MGTAB, Twibot-20, and Cresci-15 datasets compared to GCN. Notably, unlike previous multi-scale GNNs such as MixHop, etc., which are linear combinations of different order GCNs, the linear combinations of fixed $K$ low-pass filters do not effectively exploit high-frequency information.

Recently proposed methods such as FAGCN, RFA-GNN, and AdaGNN effectively utilize high-frequency information within the graph, exhibiting superior detection performance compared to previous GNN approaches. Our proposed MSGS, however, surpasses FAGCN, RFA-GNN, and AdaGNN in detection performance by flexibly adjusting frequency responses based on different datasets, thereby achieving the best results.

\subsection{Different Training Set Partition}
\label{sec:Evaluation_real}

\begin{table*}[h]
\caption{Performance of MSGS and baselines with different scales of training data. The bset results is highlighted in bold.}
\begin{center}
\setlength{\columnsep}{1pt}%
\begin{adjustbox}{width=0.925\linewidth}
\begin{tabular}{|c|l|c|c|c|c|c|c|c|}
\hline
\multicolumn{1}{|c|}{Dataset} & \multicolumn{1}{c|}{Method} & \multicolumn{1}{c|}{0.10} & \multicolumn{1}{c|}{0.15} & \multicolumn{1}{c|}{0.20} & \multicolumn{1}{c|}{0.25} & \multicolumn{1}{c|}{0.30} & \multicolumn{1}{c|}{0.35} & \multicolumn{1}{c|}{0.40} \\
\hline
\multicolumn{1}{|c|}{\multirow{5}[1]{*}{MGTAB}}
& GCN   & 84.51$\pm$1.04 & 84.93$\pm$0.80 & 85.44$\pm$0.60 & 85.39$\pm$0.56
        & 85.35$\pm$1.09 & 85.65$\pm$0.35 & 85.89$\pm$0.53 \\
\cline{2-9}
& GAT   & 84.56$\pm$0.98 & 84.99$\pm$0.60 & 85.49$\pm$0.38 & 85.84$\pm$0.57
        & 85.77$\pm$0.37 & 86.10$\pm$0.59 & 86.31$\pm$0.63 \\
\cline{2-9}
& FAGCN & 86.05$\pm$0.75 & 86.13$\pm$0.11 & 86.64$\pm$0.53 & 86.75$\pm$0.66
        & 86.82$\pm$0.55 & 87.03$\pm$0.58 & 87.63$\pm$0.76 \\
\cline{2-9}
& RFA-GNN & 86.19$\pm$0.07 & 86.23$\pm$0.28 & 86.86$\pm$0.50 & 86.99$\pm$0.98
        & 87.23$\pm$1.20 & 87.65$\pm$0.60   & 87.98$\pm$0.59 \\
\cline{2-9}
& MSGS  & \textbf{87.76$\pm$0.86} & \textbf{88.26$\pm$0.42} & \textbf{88.70$\pm$0.43} & \textbf{89.54$\pm$0.45}
        & \textbf{90.73$\pm$0.31} & \textbf{92.22$\pm$0.48} & \textbf{93.53$\pm$0.72} \\
\hline
\multicolumn{1}{|c|}{\multirow{5}[1]{*}{Twibot-20}}
& GCN   & 68.05$\pm$0.29 & 69.86$\pm$0.58 & 70.70$\pm$0.53 & 71.17$\pm$0.50
        & 72.32$\pm$0.26 & 72.69$\pm$0.45 & 73.20$\pm$0.36 \\
\cline{2-9}
& GAT   & 72.43$\pm$0.57 & 73.66$\pm$0.13 & 74.38$\pm$0.81 & 75.62$\pm$0.37
        & 76.77$\pm$0.19 & 76.92$\pm$0.44 & 77.10$\pm$0.61 \\
\cline{2-9}
& FAGCN & 78.79$\pm$0.35 & 80.76$\pm$0.28 & 81.71$\pm$0.24 & 82.32$\pm$0.33
        & 83.23$\pm$0.16 & 83.34$\pm$0.48 & 83.51$\pm$0.32 \\
\cline{2-9}
& RFA-GNN & 80.45$\pm$1.09 & 82.23$\pm$0.62 & 83.12$\pm$0.20 & 83.45$\pm$0.33
          & 83.74$\pm$0.25 & 84.02$\pm$0.45 & 84.17$\pm$0.64 \\
\cline{2-9}
& MSGS  & \textbf{82.21$\pm$0.75} & \textbf{83.52$\pm$0.37} & \textbf{83.78$\pm$0.33} & \textbf{84.56$\pm$0.49}
        & \textbf{84.77$\pm$0.88} & \textbf{84.96$\pm$0.93} & \textbf{85.41$\pm$0.68} \\
\hline
\multicolumn{1}{|c|}{\multirow{5}[1]{*}{Cresci-15}}
& GCN   & 95.27$\pm$0.71 & 95.60$\pm$0.88 & 95.73$\pm$0.71 & 96.28$\pm$0.41
        & 96.92$\pm$0.59 & 96.97$\pm$0.82 & 97.10$\pm$0.50 \\
\cline{2-9}
& GAT   & 95.59$\pm$0.45 & 96.12$\pm$0.25 & 96.08$\pm$0.30 & 96.58$\pm$0.37
        & 97.01$\pm$0.36 & 97.07$\pm$0.28 & 97.18$\pm$0.31 \\
\cline{2-9}
& FAGCN & 95.76$\pm$0.56 & 96.13$\pm$0.54 & 96.42$\pm$0.98 & 96.57$\pm$0.66
        & 96.93$\pm$0.81 & 96.98$\pm$0.83 & 97.24$\pm$0.72 \\
\cline{2-9}
& RFA-GNN & 95.94$\pm$0.36 & 96.32$\pm$0.42 & 96.45$\pm$0.72 & 96.69$\pm$0.83
          & 96.98$\pm$0.49 & 97.13$\pm$0.60 & 97.20$\pm$0.35 \\
\cline{2-9}
& MSGS  & \textbf{96.47$\pm$0.09} & \textbf{96.76$\pm$0.75} & \textbf{96.85$\pm$0.21} & \textbf{97.19$\pm$0.28}
        & \textbf{97.40$\pm$0.38} & \textbf{97.64$\pm$0.21} & \textbf{97.98$\pm$0.13} \\
\hline
\end{tabular}
\end{adjustbox}
\end{center}
\label{tb:Training_set}
\end{table*}

To further evaluate the performance enhancement of our approach, we conducted a comprehensive comparison between MSGS and other GNNs across various training sets. Specifically, we employed a validation set with a scale of 0.1 and a test set of 0.5. By varying the training set from 0.1 to 0.4, the results are presented in Table~\ref{tb:Training_set}. Notably, MSGS surpasses the baseline models by a significant margin across all social bot detection datasets, regardless of the training set. On the MGTAB, Twibot-20, and Cresci-15 datasets, MSGS achieves an average accuracy improvement of 5.55\%, 1.24\%, and 0.78\% over the best-performing baseline, respectively.

\subsection{Ablation Analysis}
\label{sec:Ablation}
In this section, we conduct a comparative analysis between MSGS and its three variants to assess the effectiveness of the designed modules. The following is a detailed description of these variations:

\begin{itemize}
\item \textbf{MSGS w/o MS} removes the multi-scale structure and solely utilizes the output from the final layer of the GNN model.
\item \textbf{MSGS w/o SAM (N)} eliminates the node-level signed-attention mechanism, setting $\alpha=1$ and $\beta=0$.
\item \textbf{MSGS w/o SAM (S)} excludes the scale-level signed-attention mechanism.

\item \textbf{MSGS} incorporates all modules within the multi-scale graph learning framework.
\end{itemize}

The second half of Table~\ref{tb:main_results} presents the performance of various variants, highlighting the roles of different modules within our proposed MSGS. Among all the variants, MSGS w/o SAM (N) exhibits the worst performance. This is because, without the node-level signed-attention mechanism, MSGS degenerates into a fixed low-pass filter, unable to effectively utilize high-frequency information. On the other hand, MSGS w/o MS removes the multi-scale structure, resulting in a significant decline in performance as it cannot leverage multi-scale representations. Conversely, MSGS w/o SAM (S), which excludes the scale-level signed-attention mechanism, demonstrates improved performance compared to MSGS w/o MS when able to utilize multi-scale features. MSGS w/o SAM (S), which averages the multi-scale features, is not as flexible as attention-based weighting. As a result, its performance is still inferior to MSGS.

\subsection{Alleviating Over-Smoothing Problem}
\label{sec:Over-smoothing}
\begin{figure}[htbp]
	\centering
    \subfloat[]{
	\begin{minipage}{0.95\linewidth}
		\centering
		\includegraphics[width=0.90\linewidth]{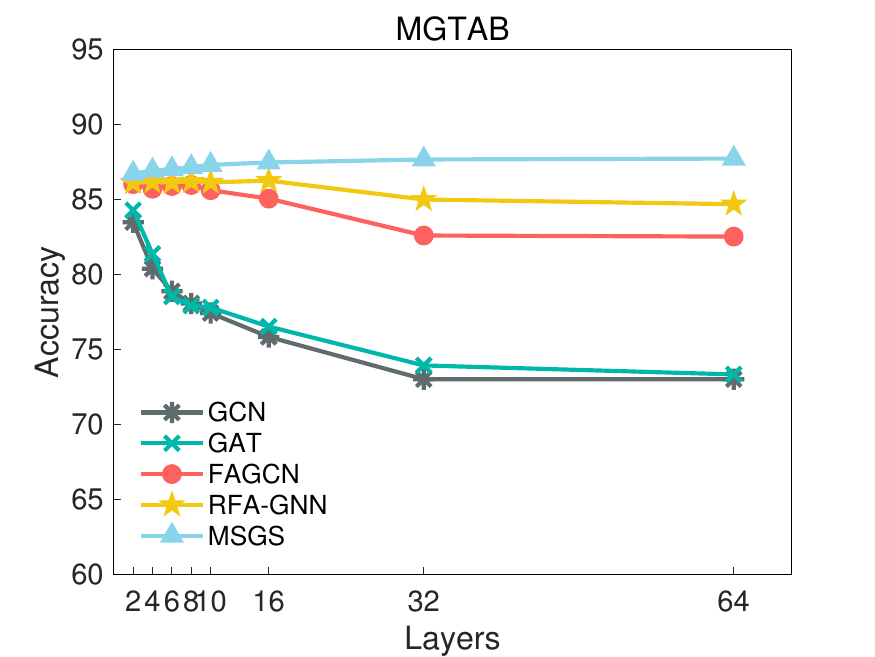}
	\end{minipage}
    }

	\subfloat[]{
	\begin{minipage}{0.95\linewidth}
		\centering
		\includegraphics[width=0.90\linewidth]{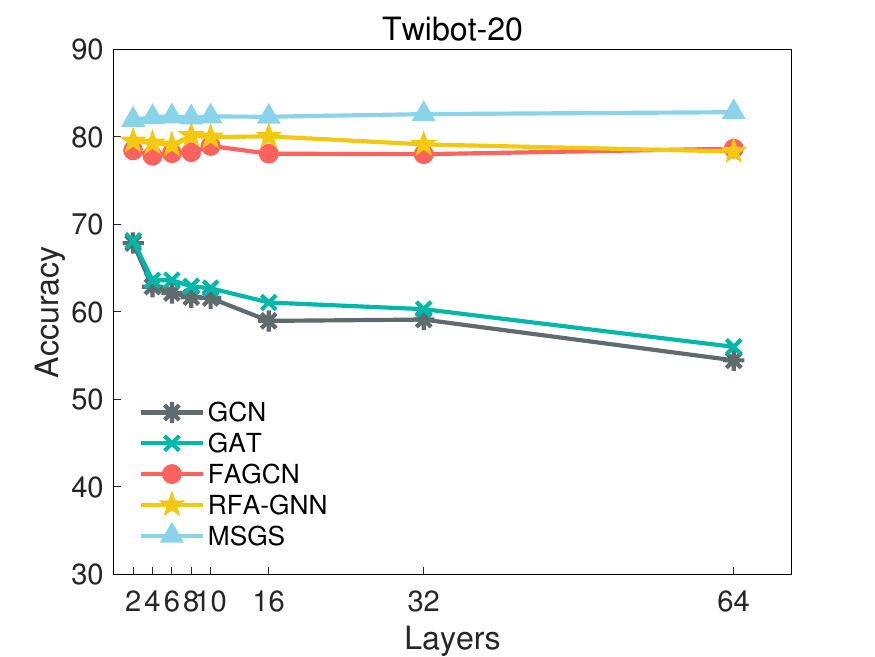}
	\end{minipage}
    }
\label{fig:GNN_layers}
\caption{The accuracy on MGTAB (a) and Twibot-20 (b) datasets with different layers.}
\end{figure}

To verify the ability of MSGS to alleviate the over-smoothing problem, we compared the performance of MSGS with GCN, FAGCN, and RFA-GCN models at different depths. We varied the number of layers in the models to $\{$2, 4, 6, 8, 10, 16, 32, 64$\}$, and the results are shown in Fig.~\ref{fig:GNN_layers}. GCN achieved the best performance at two layers, but its performance gradually decreased as the number of layers increased, demonstrating that a too-deep structure can cause severe over-smoothing in GCN models. FAGCN, RFA-GCN, and our proposed MSGS all achieved significantly higher accuracy than GCN, especially when the models had a deeper layer configuration.

GAT added an attention mechanism to the neighborhood aggregation process based on GCN, and performed better than GCN at different layer configurations. The over-smoothing problem can be slightly alleviated by the attention mechanism.
FAGCN significantly outperformed GCN at different layer configurations, indicating that utilizing high-frequency information can alleviate the negative impact of over-smoothing on the model. Compared to FAGCN, the RFA-GCN model increased the range of graph filter adjustment and consistently outperformed FAGCN. Although both FAGCN and RFA-GCN can utilize high-frequency information to alleviate the over-smoothing problem, their detection accuracy slightly decreases when the model's depth is continuously increased. Our proposed MSGS, on the other hand, not only avoids over-smoothing as the number of layers increases but also improves classification performance.

\subsection{Visualization of Edge Coefficients}
\label{sec:Coefficients}
We visualize the coefficient $\beta^{(k)}$, extracted from the last layer of MSGS to verify whether MSGS can learn different edge coefficients for different datasets. We categorize the edges in the social network graph into intra-class and inter-class based on the labels of the connected nodes. In terms of the spatial domain, low-frequency information in the graph originates from intra-class edges, while high-frequency information originates from inter-class edges.

\begin{figure*}[htbp]
	\centering
    \subfloat[]{
	\begin{minipage}{0.32\linewidth}
		\centering
		\includegraphics[width=0.9\linewidth]{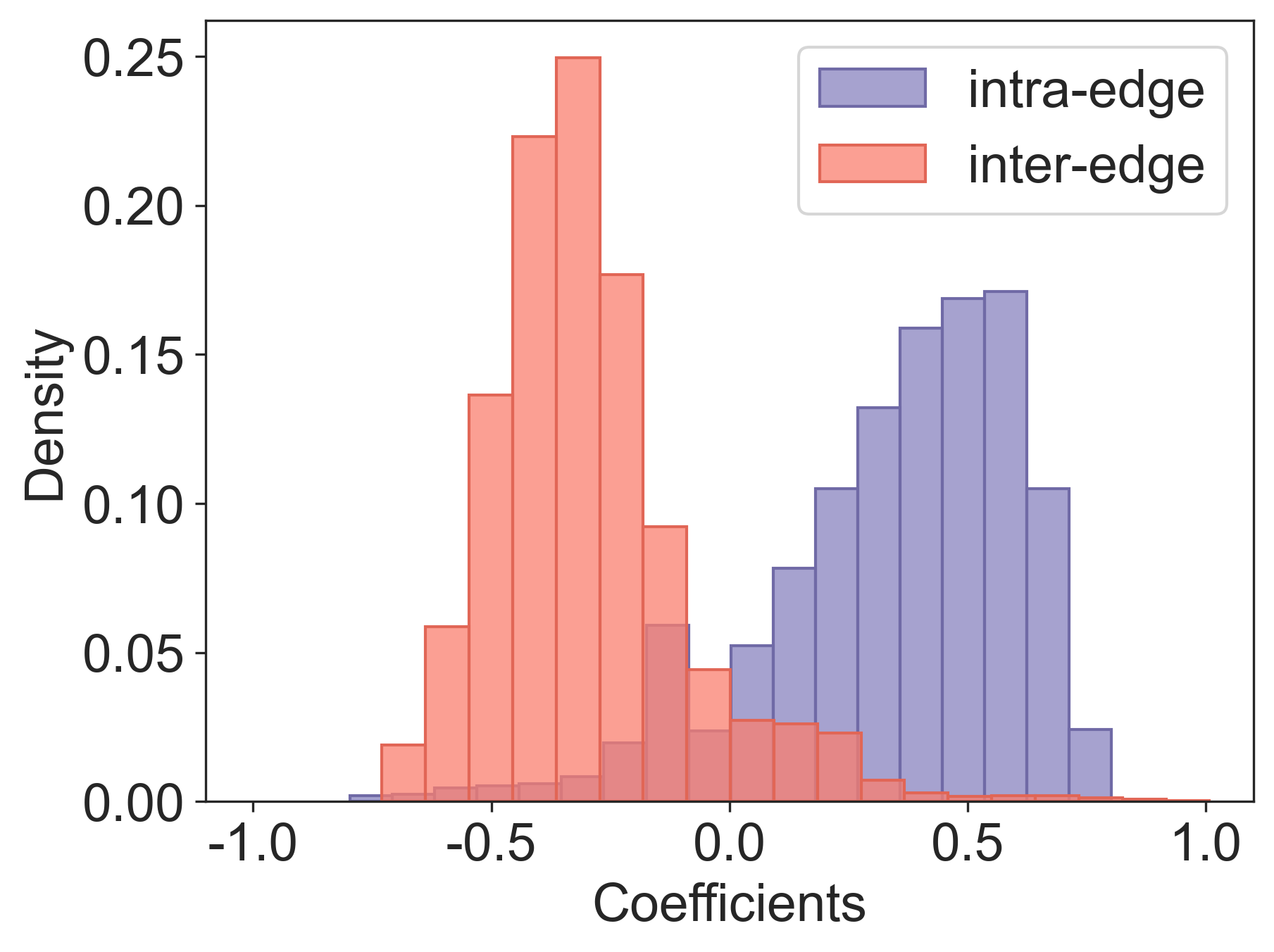}
	\end{minipage}
    }
    \subfloat[]{
	\begin{minipage}{0.32\linewidth}
		\centering
		\includegraphics[width=0.9\linewidth]{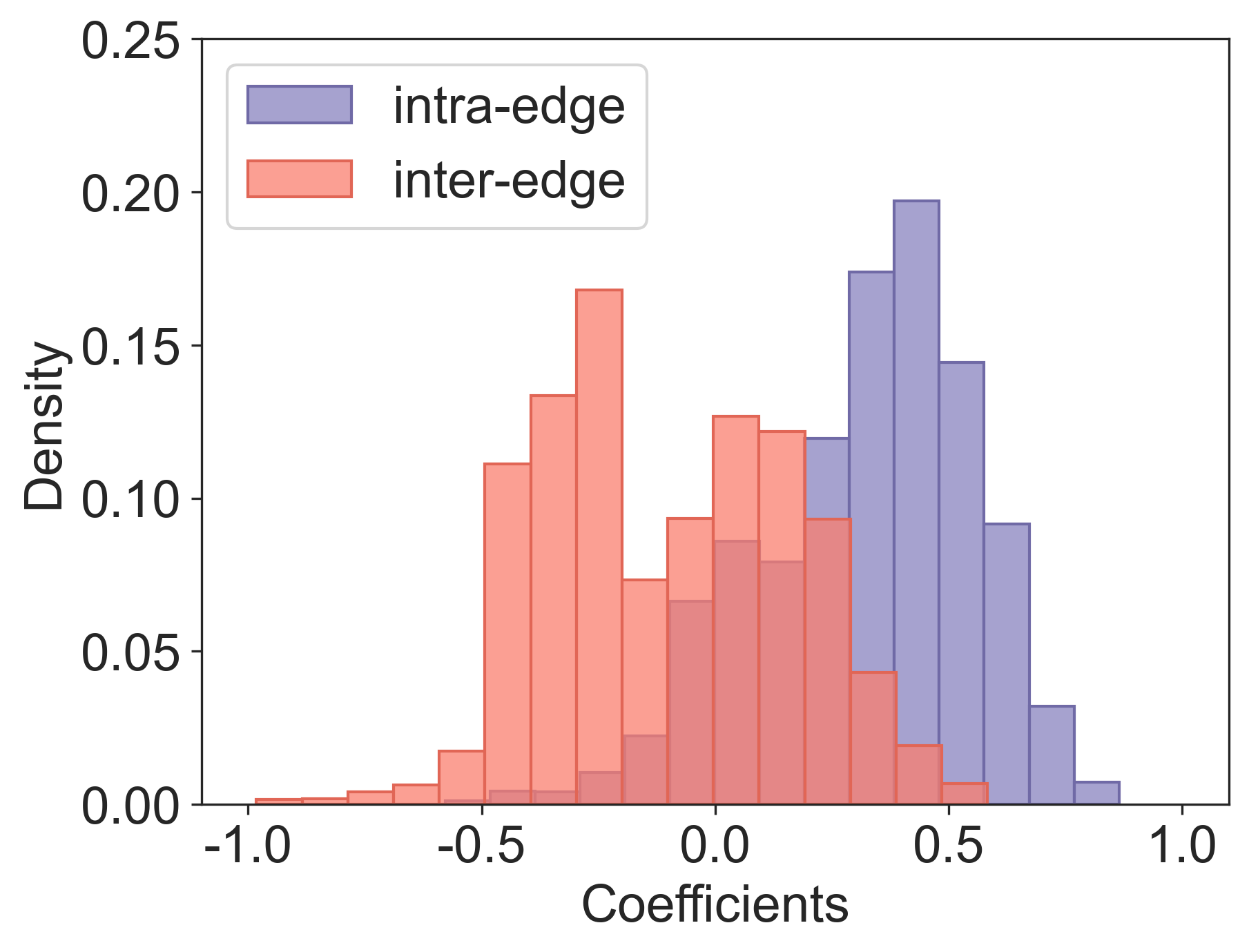}
	\end{minipage}
    }
    \subfloat[]{
	\begin{minipage}{0.32\linewidth}
		\centering
		\includegraphics[width=0.9\linewidth]{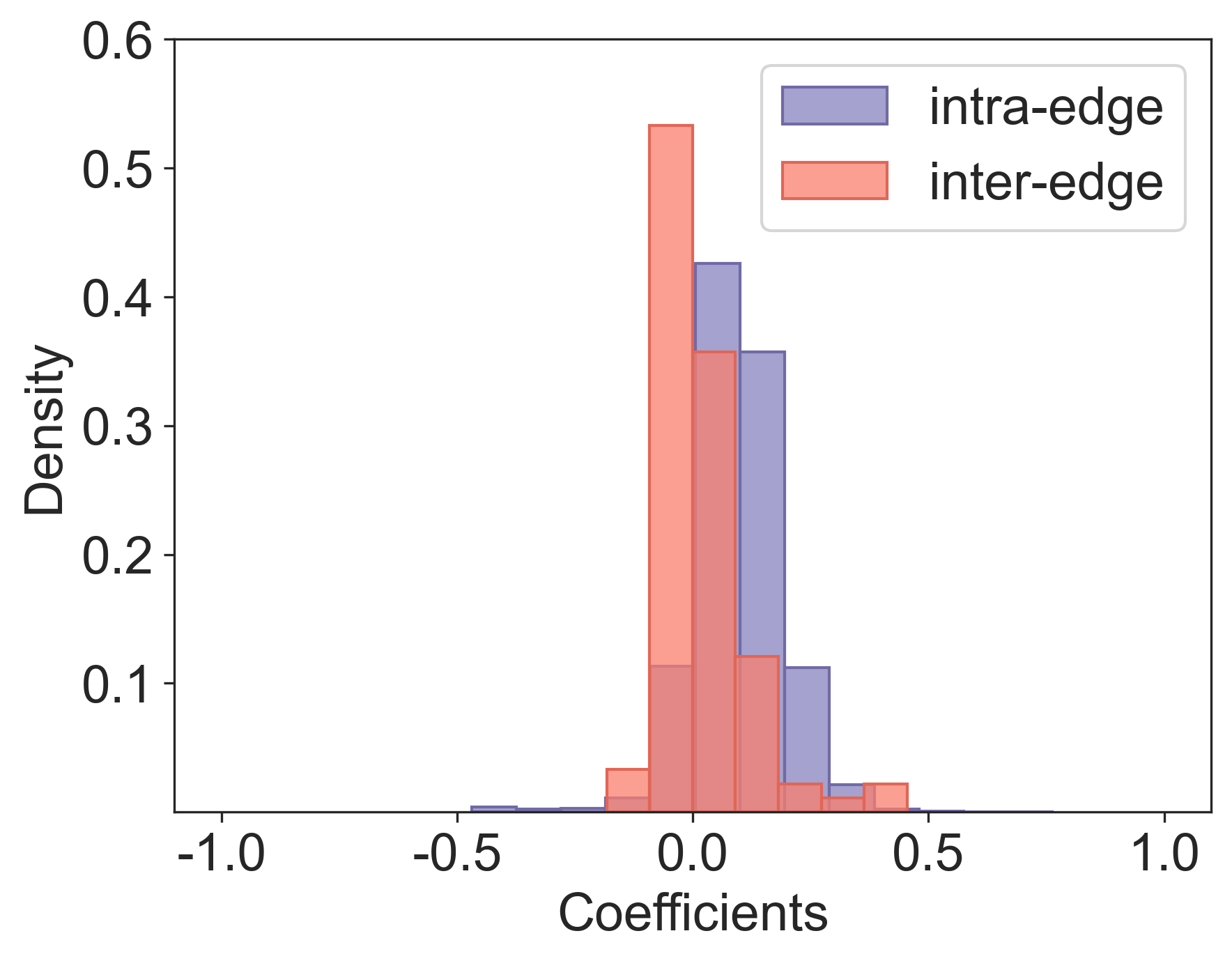}
	\end{minipage}
    }
\label{fig:coefficients}
\caption{Visualization of the mean frequency coefficients on MGTAB (a), Twibot-20 (b) and Cresci-15 (c) datasets.}
\end{figure*}

In GCN, all edges are assigned positive weights, assuming that nodes share similar features with their normal neighbors. However, high-frequency information also plays an essential role in bot detection, and anomalous nodes may connect with normal nodes, forming inter-class edges. Aggregating the neighborhood through intra-class edges can enhance the original features of the nodes, while aggregation through inter-class edges may destroy them. Our proposed MSGS allows for adaptive learning of edge weights. As shown in Fig.~\ref{fig:coefficients}, most inter-class edges have negative weights, while most intra-class edges have positive weights. This effectively utilizes high-frequency information. This allows MSGS to prioritize and leverage the important high-frequency components in the graph, enhancing its ability to capture fine-grained details and subtle patterns in the data. By incorporating this signed-attention mechanism, MSGS can effectively utilize low-frequency and high-frequency information for social bot detection.

\subsection{Visualization of Graph Filters}
\label{sec:Filter}

\begin{figure*}[htbp]
	\centering
    \subfloat[]{
	\begin{minipage}{0.32\linewidth}
		\centering
		\includegraphics[width=0.9\linewidth]{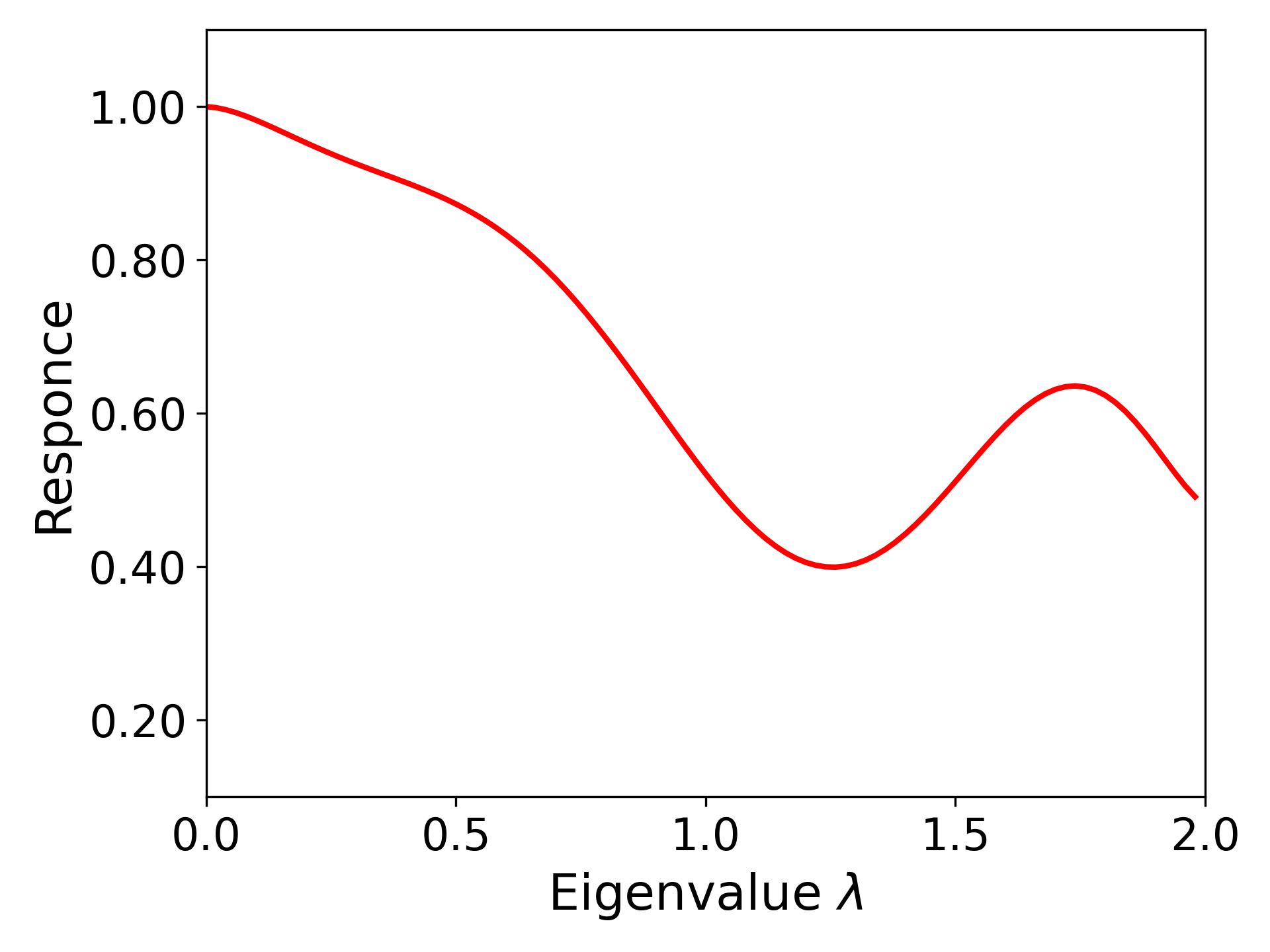}
	\end{minipage}
    }
    \subfloat[]{
	\begin{minipage}{0.32\linewidth}
		\centering
		\includegraphics[width=0.9\linewidth]{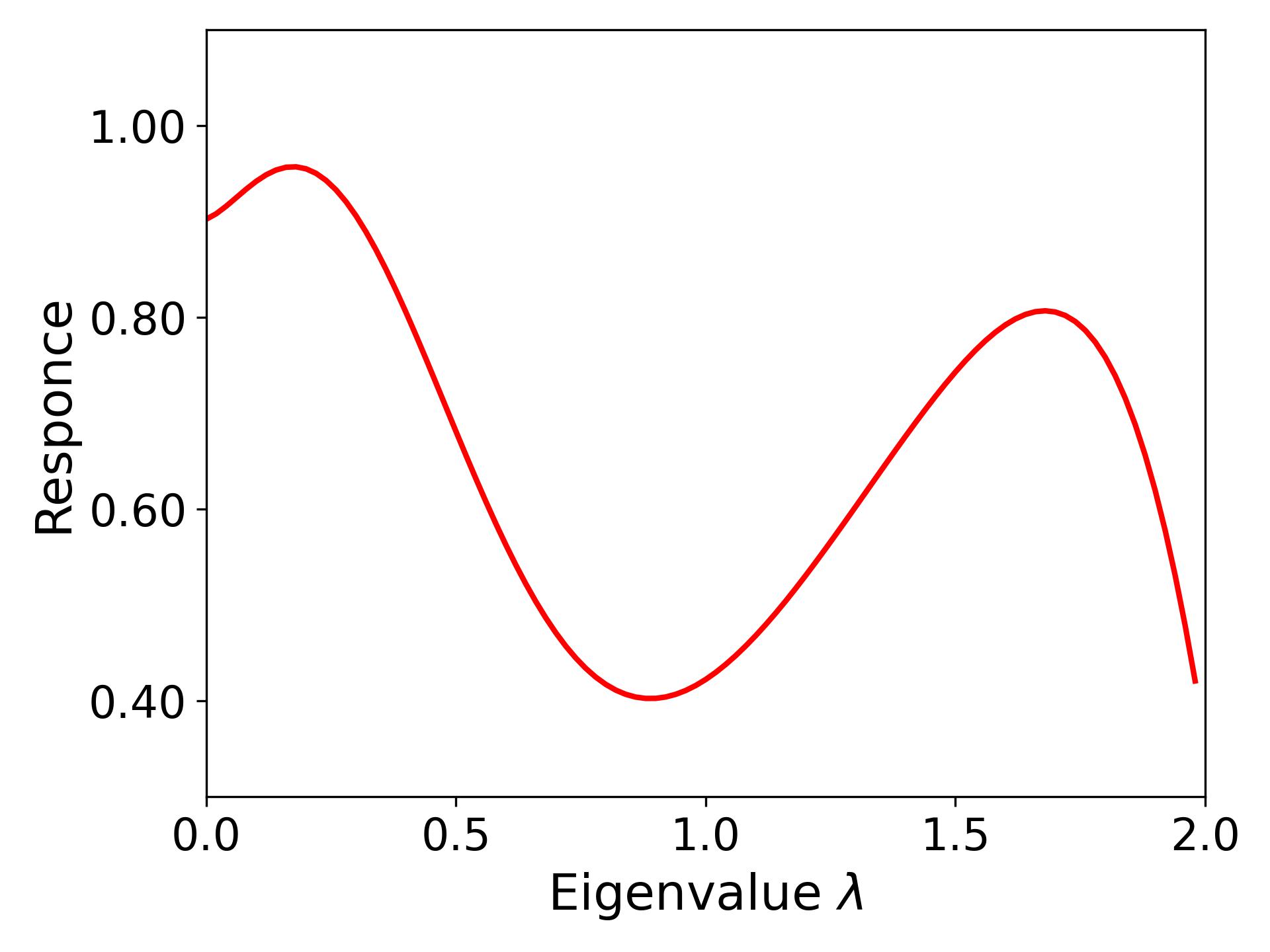}
	\end{minipage}
    }
    \subfloat[]{
	\begin{minipage}{0.32\linewidth}
		\centering
		\includegraphics[width=0.9\linewidth]{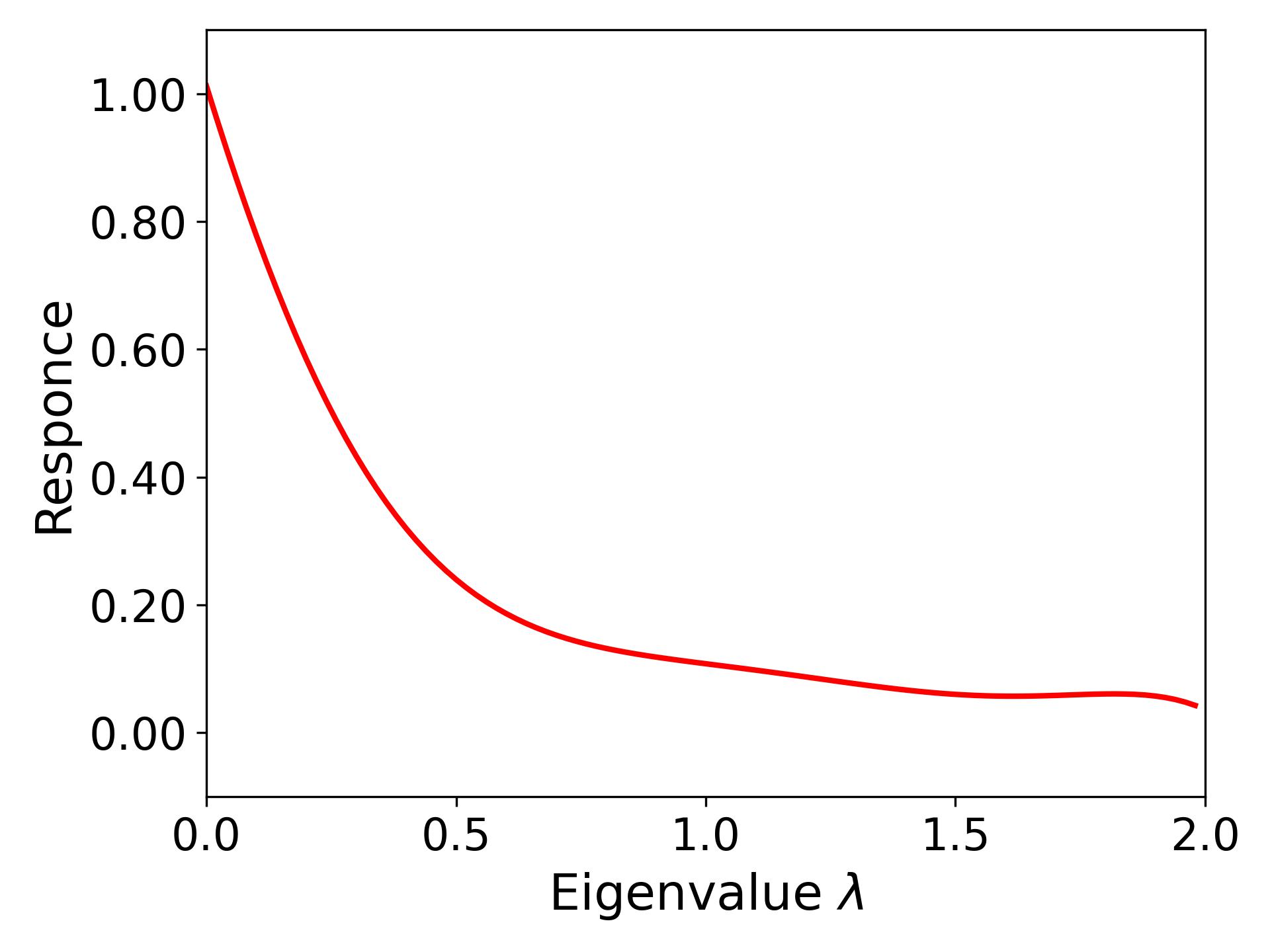}
	\end{minipage}
    }
\label{fig:filter}
\caption{MSGS’s equivalent graph filters on MGTAB (a), Twibot-20 (b) and Cresci-15 (c) datasets.}
\end{figure*}

We have generated an approximate filter for MSGS on various datasets to gain a more profound understanding of our model. Fig.~\ref{fig:filter} illustrates that our approach can effectively learn appropriate filtering patterns from the data. In the cases of MGTAB and Twibot-20, MSGS pays attention to low-frequency and high-frequency information. However, Twibot-20 exhibits more high-frequency information than MGTAB, resulting in stronger responses for the obtained graph filters in the high-frequency domain. Conversely, for Cresci-15, MSGS primarily focuses on utilizing low-frequency information for classification. Therefore, on Cresci-15, MSGS behaves similarly to previous low-frequency filtered GNNS. This explains why MSGS did not improve significantly on the Cresci-15 dataset.

\section{Related work}
\subsection{Social Bot Detection}
Social bot detection methods can be broadly categorized into feature-based and graph-based approaches. Feature-based methods~\cite{article03,article28,article32,article33} rely on feature engineering to design or extract effective detection features and then employ machine learning classifiers for classification. Early research~\cite{article03,article10,article32} utilized features such as the number of followers and friends and the number of tweets for detection. Subsequent work incorporated account posting content features to improve detection effectiveness further~\cite{article30,article31,article33}. However, feature-based methods fail to leverage the interaction relationships between users.

Graph neural networks have recently been applied to social bot detection with promising results. Compared to feature-based methods, graph neural networks effectively utilize user interaction features, such as follow and friend relationships~\cite{article12}. Graph neural network-based account detection methods~\cite{article13,article15,article16} first construct a social relationship graph and then transform the problem of detecting bot accounts into a node classification problem. Feng et al.~\cite{article16} constructed a social relationship graph using friend and follower relationships, extracted tweet features, description features, and identity field features of the accounts, and then performed node classification using RGCN. OS3-GNN~\cite{article15} is a graph neural network framework that addresses the issue of class imbalance in social bot detection by generating minority class nodes in the feature space, thereby alleviating the imbalance between human and bot accounts. Shi et al.~\cite{article36} proposed a graph ensemble learning method that combines random forest~\cite{article25} with GNN for social bot detection.

\subsection{Graph Neural Networks}
Graph Neural Networks are neural networks designed for processing graph data. Unlike traditional methods, GNNs enable information exchange and aggregation among nodes by defining message passing on nodes and edges. Compared to traditional graph embedding methods such as DeepWalk~\cite{article26} and node2vec~\cite{article27}, GNNs have the capability to learn richer and more advanced node representations through multi-layer stacking and information propagation mechanisms. GNNs effectively capture relationships and global structures among nodes in graphs, making them suitable for various domains such as social network analysis, recommendation systems, and molecular graph analysis~\cite{article17}.

Inspired by graph spectral theory, a learnable graph convolution operation was introduced in the Fourier domain~\cite{article24}. GCN~\cite{article17} simplified the convolution operation using a linear filter, becoming the most prevalent approach. GAT~\cite{article19} introduced an attention mechanism to weigh the feature sum of neighboring nodes based on GCN. APPNP~\cite{article14} utilizes Personalized PageRank~\cite{article11}, constructing a low-pass filter with distinct concentration properties compared to GCN. Several algorithms~\cite{article08,article20,article21,article23} have contributed to the improvement of GCN and enhanced the performance of GNNs.

Existing spectral GNNs primarily employ fixed filters for the convolution operation, which can lead to over-smoothing issues due to the lack of learnability~\cite{article06}. Recently, the spectral analysis of GNNs has garnered significant interest for its valuable insights into the interpretability and expressive power of GNNs~\cite{article06,article07}. RFGCN~\cite{article04} has attempted to demonstrate that most GNNs are restricted to low-pass filters and have argued for the necessity of high-pass and band-pass filters. RFA-GNN~\cite{article05} further extends the adjustment scope of RFGCN~\cite{article04}, enabling better utilization of high-frequency information. These models enhance the expressive capacity of GNNs and enable adaptive adjustments of the frequency response of graph filters. However, their adjustment space needs to be improved. In this regard, we propose MSGS, which further expands the frequency domain adjustment space.

\section{Conclusion}
This paper introduces a novel social bot detection method called Multi-scale Graph Neural Network with Signed-Attention (MSGS). By incorporating multi-scale architecture and the signed attention mechanism, we construct an adaptive graph filter that can adjust the frequency response of the detection model based on different data, effectively utilizing both low-frequency and high-frequency information. Through the theoretical analysis from the frequency domain perspective, we have proved that MSGS expands the frequency domain adjustment space compared to existing graph filters. Moreover, MSGS addresses the over-smoothing problem commonly observed in existing GNN models. It exhibits exceptional performance, even in deep structures. Extensive experiments demonstrate that MSGS consistently outperforms state-of-the-art GNN baselines on social bot detection benchmark datasets.

\section*{Acknowledgment}
This work was supported by the National Key Research and Development Project of China (Grant No. 2020YFC1522002).

{\appendix[Proof of Theorem in paper]
\noindent \textbf{Proof of Theorem 1.}
The Fourier transform of $f$ can be expressed as: $\mathcal{F}\{f\}(v)=\int_{\mathbb{R}} f(x) e^{-2 \pi i x \cdot v} d x$. The inverse transformation can be expressed as: $\mathcal{F}^{-1}\{f\}(x)=\int_{\mathbb{R}} f(v) e^{2 \pi i x \cdot v} d v$. We define $h$ to be the convolution of $f$ and $g$, then $h(z)=\int_{\mathbb{R}} f(x) g(z-x) d x$. Taking the Fourier transform of $h$, we get:

\begin{equation}
\begin{split}
\label{equ:FT}
\mathcal{F}\{f * g\}(v) & \mathcal{F}\{h\}(v) \\
& = \int_{\mathbb{R}} h(z) e^{-2 \pi i z \cdot v} d z \\
& = \int_{\mathbb{R}} \int_{\mathbb{R}} f(x) g(z-x) e^{-2 \pi i z \cdot v} dx dz \\
& = \int_{\mathbb{R}} f(x)\left(\int_{\mathbb{R}} g(z-x) e^{-2 \pi i z \cdot v} dz\right) dx.
\end{split}
\end{equation}

We substitute $y=z-x$ and $dy=dz$ into Equ.~(\ref{equ:FT}):

\begin{equation}
\begin{split}
\label{equ:FT2}
\mathcal{F}\{f*g\}(v) & = \int_{\mathbb{R}}f(x)\left(\int_{\mathbb{R}}g(y) e^{-2 \pi i(y+x) \cdot v} dy\right) dx \\
 & = \int_{\mathbb{R}} f(x) e^{-2 \pi i x v v}\left(\int_{\mathbb{R}} g(y) e^{-2 \pi i v v v} d y\right) d x \\
 & = \int_{\mathbb{R}} f(x) e^{-2 \pi i x v v} d x \int_{\mathbb{R}} g(y) e^{-2 \pi i v v v} d y \\
 & = \mathcal{F}\{f\}(v) \cdot \mathcal{F}\{g\}(v)
\end{split}
\end{equation}

Taking the inverse Fourier transform of both sides of Equ.~(\ref{equ:FT2}), we get: $f * g=\mathcal{F}^{-1}\{\mathcal{F}\{f\} \cdot \mathcal{F}\{g\}\}$.

\noindent \textbf{Proof of Theorem 2.} For GCN, the symmetric Laplacian matrix is:

\begin{equation}
\mathbf{L}_{sym}=\mathbf{I}_{N}-\mathbf{D}^{-\frac{1}{2}} \mathbf{A} \mathbf{D}^{-\frac{1}{2}}=\mathbf{U} \mathbf{\Lambda} \mathbf{U}^{T}=\sum_{i=1}^{N} \lambda_{i} \mathbf{u}_{i} \mathbf{u}_{i}^{T},
\end{equation}

\noindent where $\lambda_{i}$ represents the eigenvalue, $1 \leq \lambda_{i} \leq N$ and $0=\lambda_{1} \leq \lambda_{2} \leq ... \leq \lambda_{N}$.

\begin{equation}
\mathbf{D}^{\frac{1}{2}} \mathbf{L}_{s y m} \mathbf{D}^{\frac{1}{2}} \mathbf{1}=(\mathbf{D}-\mathbf{A}) \mathbf{1}=\mathbf{0},
\end{equation}

\noindent where \textbf{1} is the vector with all 1 elements, and multiply both sides by the inverse of $\mathbf{D}^{\frac{1}{2}}$ to get $\mathbf{L}_{sym} \mathbf{D}^{\frac{1}{2}} \mathbf{1}=\mathbf{0}$.

So $\mathbf{L}_{sym}$ has an eigenvalue of 0 and the corresponding eigenvector $\mathbf{D}^{\frac{1}{2}} \mathbf{1}$, and the largest eigenvalue of $\mathbf{L}_{sym}$ is the upper bound of the Rayleigh quotient:

\begin{equation}
\lambda_{N}=\sup _{\mathbf{g}} \frac{\mathbf{g}^{T} \mathbf{L}_{s y m} \mathbf{g}}{\mathbf{g}^{T} \mathbf{g}},
\end{equation}

where $\mathbf{g}$ is a nonzero vector. let $\mathbf{f}=\mathbf{D}^{-\frac{1}{2}} \mathbf{g}$, then we have

\begin{equation}
\begin{split}
\frac{\mathbf{f}^{T} L f}{\left(\mathbf{D}^{\frac{1}{2}} \mathbf{f}\right)^{T}\left(\mathbf{D}^{\frac{1}{2}} \mathbf{f}\right)} &= \frac{\sum_{(u, v) \in E}\left(\mathbf{f}_{u}-\mathbf{f}_{v}\right)^{2}}{\sum_{v \in V} f_{v}^{2} d_{v}} \\
& \leq \frac{\sum_{(u, v) \in E}\left(2 f_{u}^{2}+2 f_{v}^{2}\right)}{\sum_{v \in V} f_{v}^{2} d_{v}} = 2.
\end{split}
\end{equation}

When the graph is a binary graph, the equality sign of the inequality holds. Since in reality, as long as the graph is not too small, it is almost impossible to be a bipartite graph, so we will not discuss the case of bipartite graph. Therefore, under the assumption that it is not a bipartite graph, the maximum eigenvalue is less than 2. Since and are both symmetric normalized Lapacian matrices of a graph, the only difference is that the graph corresponding to the former is added with a self-ring, so the eigenvalues of the former are also in the range $[0,2)$.

For the formula X, ignore the activation function we can get: $\mathbf{H}^{(l)}=\hat{\mathbf{A}} \mathbf{H}^{(l-1)} \mathbf{W}^{(l)}$. Since $\hat{\mathbf{A}}=\mathbf{I}_{N}-\tilde{\mathbf{L}}_{sym}$,

\begin{equation}
\begin{split}
\hat{\mathbf{A}}^{K}&=\left(I_{N}-\tilde{L}_{s y m}\right)^{K}=\left(\mathbf{I}_{N}-\mathbf{U} \mathbf{\Lambda} \mathbf{U}^{T}\right)^{K}\\
&=\mathbf{U}\left(\mathbf{I}_{N}-\boldsymbol{\Lambda}\right)^{K} \mathbf{U}^{T}=\sum_{i=1}^{N}\left(1-\lambda_{i}\right)^{K} \mathbf{u}_{i} \mathbf{u}_{i}^{T}.
\end{split}
\end{equation}

According to the range of eigenvalues proved above, the convergence state of $\hat{\mathbf{A}}^{K}$ can be obtained:

\begin{equation}
\lim _{K \rightarrow+\infty} \hat{\mathbf{A}}^{K}=\mathbf{u}_{1} \mathbf{u}_{1}^{T}, \quad \mathbf{u}_{1}=\frac{\mathbf{D}^{\frac{1}{2}} \mathbf{1}}{\sqrt{M+N}},
\end{equation}

\noindent where $M$ and $N$ represent the number of edges and nodes, respectively,

\begin{equation}
\lim _{K \rightarrow \infty} \hat{\mathbf{A}}^{K} \mathbf{x}=C \times\left[\begin{array}{c}
\sqrt{d_{1}+1} \\
\sqrt{d_{2}+1} \\
\vdots \\
\sqrt{d_{N}+1},
\end{array}\right]
\end{equation}

\noindent where $C$ is a constant, $C=\frac{1}{M+N} \sum_{j=1}^{N}(\sqrt{d_{j}+1} x_{j})$. Therefore, when the number of layers $K$ is large, the input graph signal has been completely smoothed off, and the remaining information is only the degree, and the graph signal is difficult to be linearly separable in Euclidean space. It leads to over smoothing. As the filter of conventional GCN variants are mainly defined over $\tilde{\mathbf{L}}_{sym}$ and satisfy the above condition at extremely deep layers, thus they often suffer from the over-smoothing problem.

}

\bibliographystyle{IEEEtran}
\bibliography{ecrc}

\section{Biography Section}

\begin{IEEEbiography}[{\includegraphics[width=1in,height=1.25in,clip,keepaspectratio]{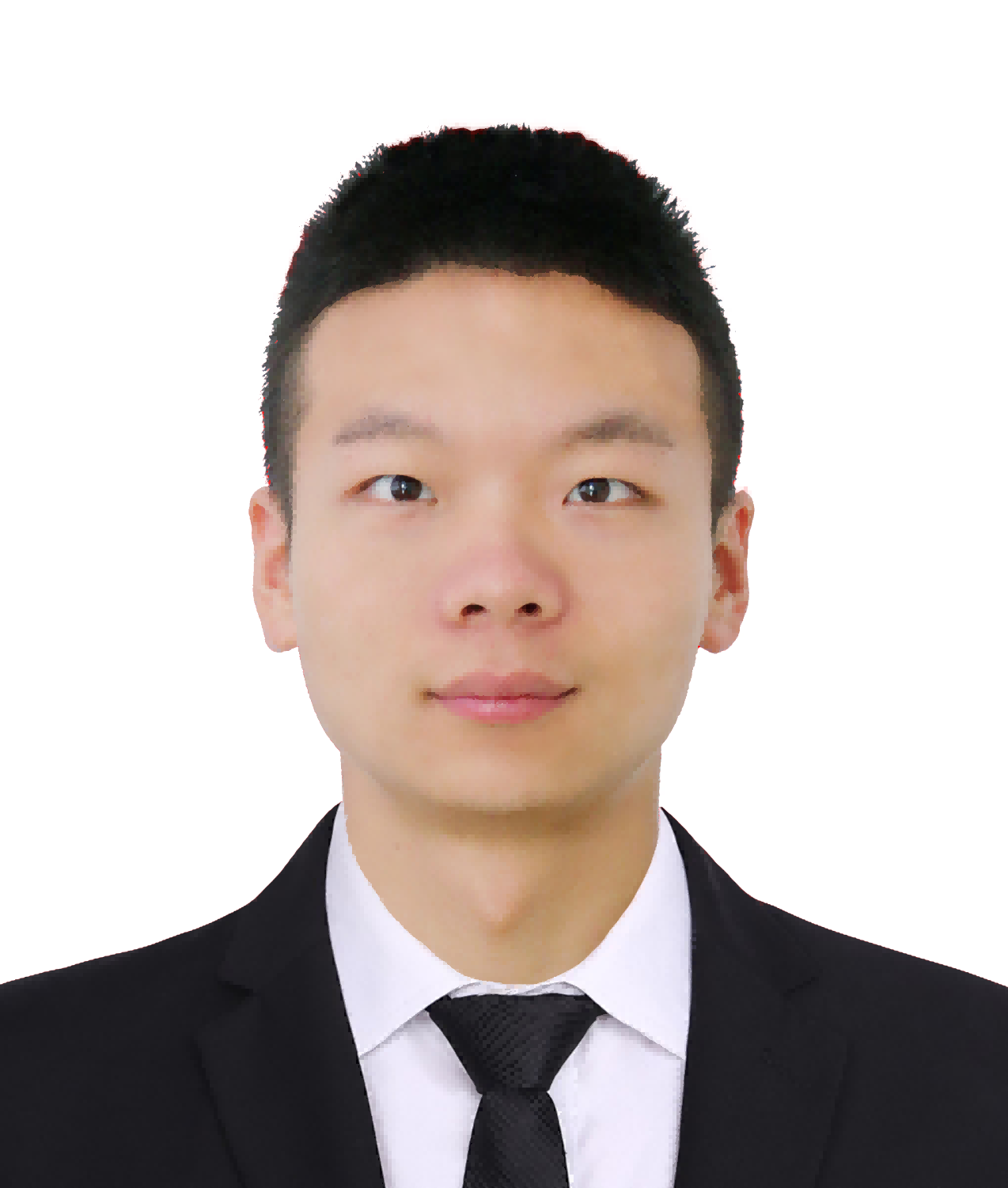}}]{Shuhao Shi}
received the B.S. degree in PLA strategy support force information engineering university, Zhengzhou, China, in 2019. He is currently pursuing his Ph.D. degree at China PLA strategy support force information engineering university. He is the author of 15 articles. His research interests include graph neural network and social media account detection.
\end{IEEEbiography}

\begin{IEEEbiography}[{\includegraphics[width=1in,height=1.25in,clip,keepaspectratio]{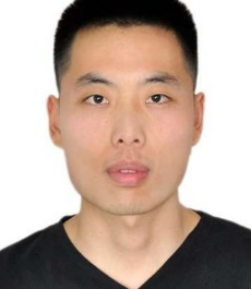}}]{Kai Qiao} received the B.S., M.S. and Ph.D. degrees in PLA strategy support force information engineering university, Zhengzhou, China, in 2014, 2017 and 2020, respectively. He is the author of 55 articles. Since 2020, he has been an Assistant Professor. His research interests include image processing, and social media account detection.
\end{IEEEbiography}

\begin{IEEEbiography}[{\includegraphics[width=1in,height=1.25in,clip,keepaspectratio]{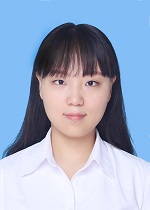}}]{Zhengyan Wang} received the B.E. degree in Central South University (CSU), Changsha, China, in 2018 and the M.E. degree in information and communication engineering from the National University of Defense Technology (NUDT), Changsha, China, in 2020. Her research interests include image processing and social media account detection.
\end{IEEEbiography}

\begin{IEEEbiography}[{\includegraphics[width=1in,height=1.25in,clip,keepaspectratio]{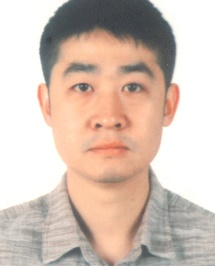}}]{Jian Chen} received the B.S., M.S. and Ph.D. degrees in PLA strategy support force information engineering university, Henan, China, in 2003, 2007 and 2013. From 2001 to 2004, he was a Research Assistant the National Digital Switching System Engineering \& Technological R \& D Centre. Since 2015, he has been an Assistant Professor. He is the author of three books, 37 articles and holds 3 patents. His research interests include graph data processing, bots detection and intelligent information processing.
\end{IEEEbiography}

\begin{IEEEbiography}[{\includegraphics[width=1in,height=1.25in,clip,keepaspectratio]{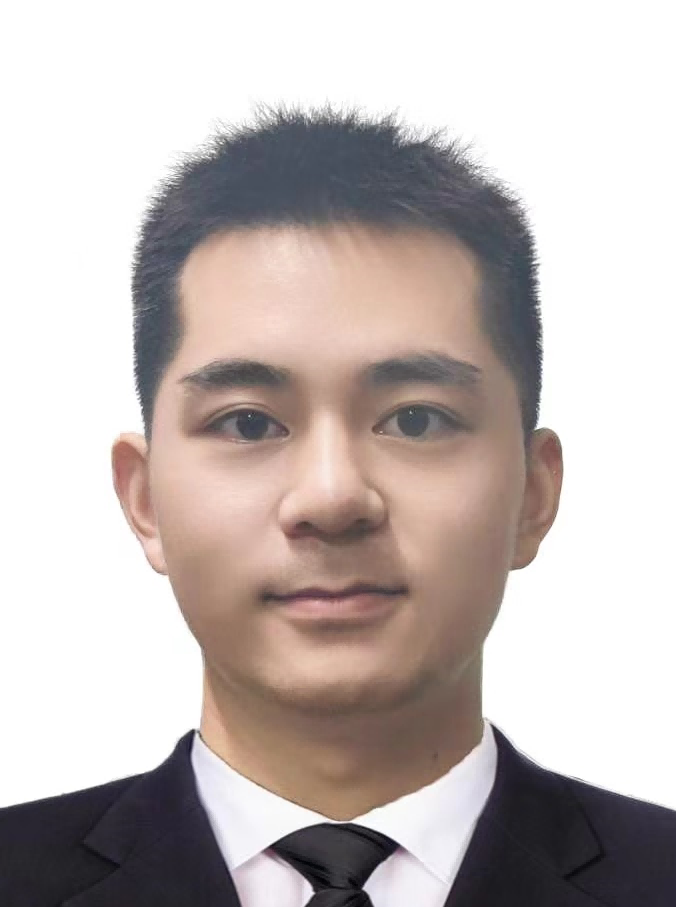}}]{Jie Yang}
received the B.S. degree in PLA strategy support force information engineering university, Zhengzhou, China, in 2021. He is currently pursuing his M.S. degree at China PLA strategy support force information engineering university. His research interests include reinforcement learning and bots detection.
\end{IEEEbiography}

\begin{IEEEbiography}[{\includegraphics[width=1in,height=1.25in,clip,keepaspectratio]{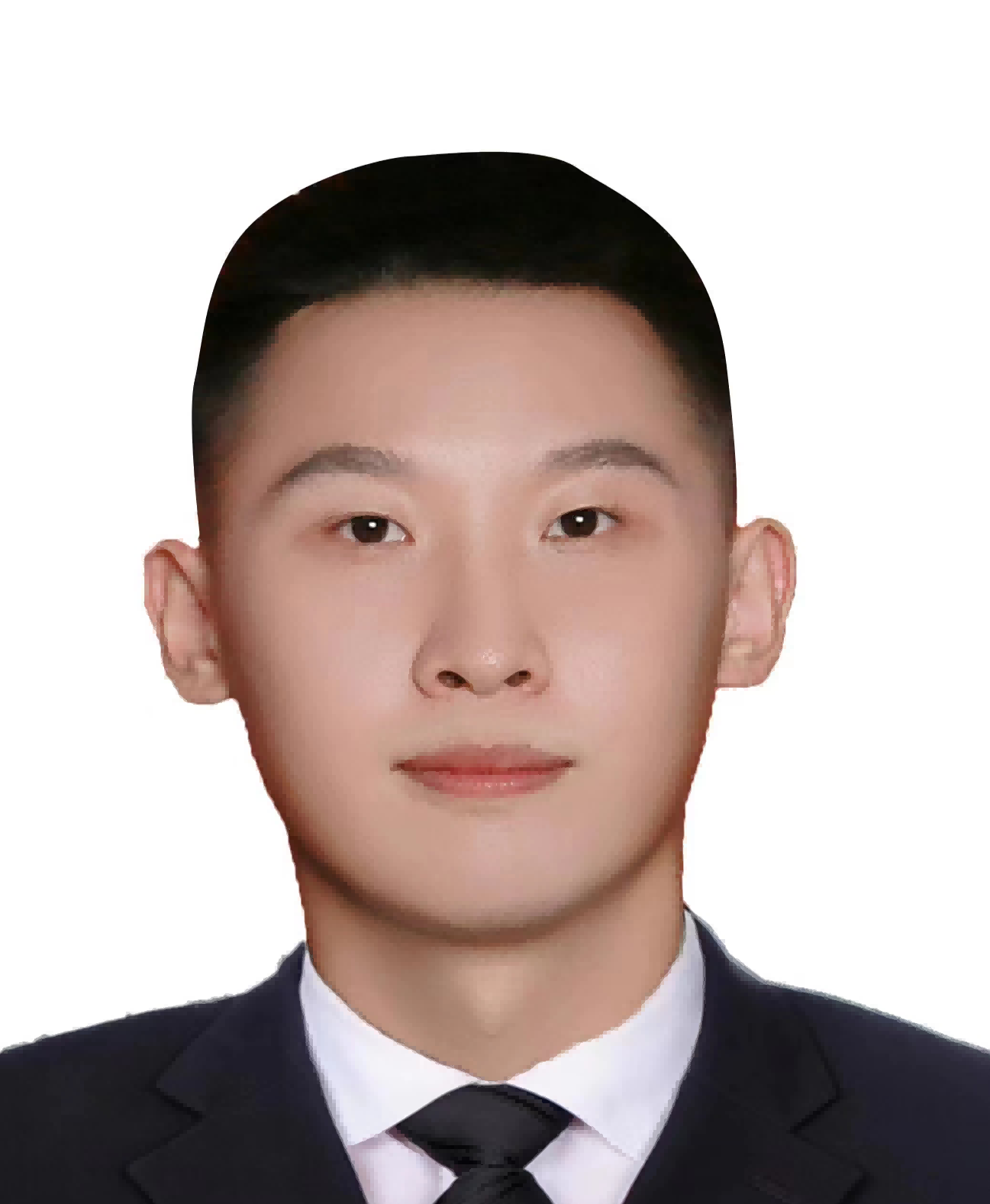}}]{Baojie Song}
received the B.S. degree in PLA strategy support force information engineering university, Zhengzhou, China, in 2021. He is currently pursuing his Ph.D. degree at China PLA strategy support force information engineering university. His research interests include semantic segmentation, object detection and natural language processing.
\end{IEEEbiography}

\begin{IEEEbiography}[{\includegraphics[width=1in,height=1.25in,clip,keepaspectratio]{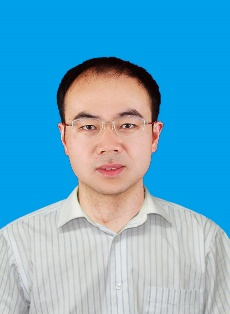}}]{Bin Yan}
received the B.S. degree in PLA strategy support force information engineering university, Zhengzhou, China, in 2002, and the Ph.D. degree in Institute of High Energy Physics Chinese Academy of Sciences, in 2006. From 2006 to 2009. He was a Research Assistant with the National Digital Switching System Engineering \& Technological R \& D Centre. From 2009 to 2015, he was an Assistant Professor with the National Digital Switching System Engineering \& Technological R \& D Centre and Henan Key Laboratory of Imaging and Intelligence Processing. Since 2015 he has been a Professor in Henan Key Laboratory of Imaging and Intelligence Processing. He is the author of three books, more than 200 articles and holds 5 patents. His current research is focused on intelligent information process.
\end{IEEEbiography}

\vspace{11pt}
\vfill

\end{document}